\pdfoutput=1

\documentclass[11pt, a4paper, table, xcdraw]{article}

\usepackage[OT2, T1]{fontenc}
\usepackage[russian, english]{babel}
\usepackage{substitutefont}

\usepackage{slashbox}

\usepackage[table,xcdraw]{xcolor}
\usepackage[]{acl}

\usepackage{booktabs}
\usepackage{times}
\usepackage{latexsym}
\usepackage{graphicx}
\usepackage{amsmath, amssymb}
\usepackage{caption}
\usepackage{subcaption}

\usepackage{placeins} 
\usepackage{stmaryrd}
\usepackage{url}
\usepackage{xr}

\usepackage[T1]{fontenc}
\usepackage{tabularx}
\usepackage[normalem]{ulem}
\useunder{\uline}{\ul}{}
\usepackage{multirow}

\usepackage[utf8]{inputenc}

\usepackage{microtype}
\usepackage{tikz}
\usetikzlibrary{positioning}

\newcommand{\oa}[1]{}

\newcommand{\as}[1]{}

\graphicspath{{graphs/}}

\newcommand{\bb}[1]{\mathbb{#1}}

\title{Semantics-aware Attention Improves Neural Machine Translation}
\author{Aviv Slobodkin \hspace{1.4cm} Leshem Choshen \hspace{1.4cm} Omri Abend \\
        School of Computer Science and Engineering\\
        The Hebrew University of Jerusalem\\
        {\tt \{aviv.slobodkin,leshem.choshen,omri.abend\}@mail.huji.ac.il}}

\date{}

\begin{document}
\maketitle
\begin{abstract}
    The integration of syntactic structures into Transformer machine translation has shown positive results, but to our knowledge, no work has attempted to do so with semantic structures. In this work we propose two novel parameter-free methods for injecting semantic information into Transformers, both rely on semantics-aware masking of (some of) the attention heads. One such method operates on the encoder, through a Scene-Aware Self-Attention (SASA) head. Another on the decoder, through a Scene-Aware Cross-Attention (SACrA) head. We show a consistent improvement over the vanilla Transformer and syntax-aware models for four language pairs.
    We further show an additional gain when using both semantic and syntactic structures in some language pairs.
\end{abstract}

\section{Introduction}

    It has long been argued that semantic representation can benefit machine translation \citep{weaver_translation_1955, BarHillel1960ThePS}.
    Moreover, RNN-based neural machine translation (NMT) has been shown to benefit from the injection of semantic structure \citep{song-etal-2019-semantic, marcheggiani-etal-2018-exploiting}. Despite these gains, to our knowledge, there have been no attempts to incorporate semantic structure into NMT Transformers \citep{vaswani2017attention}. We address this gap, focusing on the main events in the text, as represented by UCCA \citep[Universal Cognitive Conceptual Annotation;][]{abend2013universal}, namely \textit{scenes}.

\begin{figure*}
     \centering
     \begin{subfigure}[b]{0.75\textwidth}
         \centering
         \includegraphics[width=\textwidth]{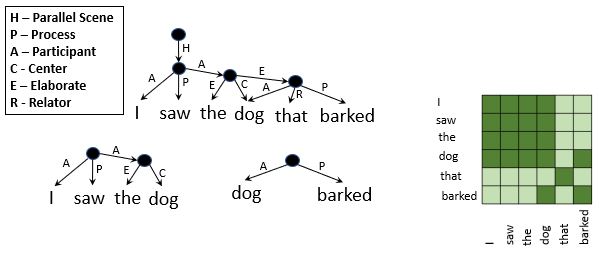}
         \caption{I saw the dog that barked.}
         \label{fig:ucca_to_mask_example1}
     \end{subfigure}
     \hfill
     \begin{subfigure}[b]{0.76\textwidth}
         \centering
         \includegraphics[width=\textwidth]{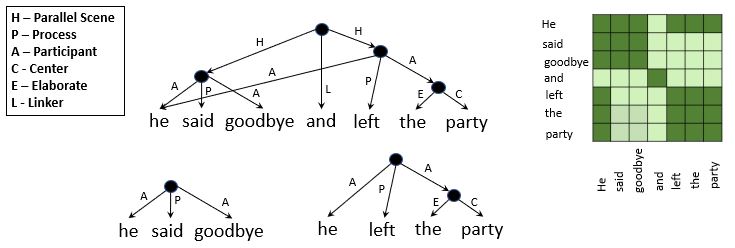}
         \caption{He said goodbye and left the party.}
         \label{fig:ucca_to_mask_example2}
     \end{subfigure}
        \caption{Examples of UCCA parse graphs of the sentences "I saw the dog that barked" (\ref{fig:ucca_to_mask_example1}) and "He said goodbye and left the party" (\ref{fig:ucca_to_mask_example2}), accompanied by their segmentation into scenes ( + corresponding UCCA sub-graphs) and equivalent Scene-Aware masks. The dark-green color in the masks represents the value '1', and the light-green color to the value '0'.}
        \label{fig:ucca_to_masks}
\end{figure*}

    UCCA is a semantic framework originating from typological and cognitive-linguistic theories \citep{dixon2009basic, dixon2010basic, dixon2012basic}. Its principal goal is to represent some of the main elements of the semantic structure of the sentence while disregarding its syntax. Formally, a UCCA representation of a passage is a directed acyclic graph where leaves correspond to the words of the sentence and nodes correspond to semantic units. The edges are labeled by the role of their endpoint in the relation corresponding to their starting point (see Fig.~\ref{fig:ucca_to_masks}). One of the motivations for using UCCA is its capability to separate the sentence into \textit{"Scenes"}, which are analogous to events (see  Fig.~\ref{fig:ucca_to_masks}). Every such Scene consists of one main relation, which can be either a Process (i.e., an action), denoted by P, or a State (i.e., continuous state), denoted by S. Scenes also contain at least one Participant (i.e., entity), denoted by A. For example, the sentence in  Fig.~\ref{fig:ucca_to_mask_example1} comprises two scenes: the first one has the Process "saw" and two Participants -- "I" and "the dog"; the second one has the Process "barked" and a single Participant -- "dog".
    
    So far, to the best of our knowledge, the only structure-aware work that integrated linguistic knowledge and graph structures into Transformers used syntactic structures \citep[][{\it inter alia}]{strubell-etal-2018-linguistically,bugliarello-okazaki-2020-enhancing, akoury-etal-2019-syntactically, sundararaman2019syntaxinfused, choshen2021transition}. The presented method builds on the method proposed by \citet{bugliarello-okazaki-2020-enhancing}, which utilized a Universal Dependencies graph \citep[UD;][]{nivre2016universal} of the source sentence to focus the encoder's attention on each token's parent, namely the token's immediate ancestor in the UD graph. Similarly, we use the UCCA graph of the source sentence to generate a scene-aware mask for the self-attention heads of the encoder. We call this method \textit{SASA} (see \S\ref{subsec:Scene-Aware Self-Attention}).


    We test our model (\S\ref{sec:Model}) on translating English into four languages. Two that are more syntactically similar to English \citep{nikolaev2020finegrained,dryer2013world}: German (En-De), Russian (En-Ru), and two that are much less so: Turkish (En-Tr) and Finnish (En-Fi). We selected these language pairs for their varied grammatical properties and the availability of reliable parallel datasets for each of them in the WMT benchmark. 
    We find consistent improvements across multiple test sets for all four cases. 
    
    In addition, we create a syntactic variant of our semantic model for better comparability. We observe that on average, our semantically aware model outperforms the syntactic models. Moreover, for the two languages less similar to English (En-Tr and En-Fi), combining both the semantic and the syntactic data results in a further gain. While improvements are often small, at times the combined version outperforms SASA and UDISCAL (our syntactic variant, see \S\ref{sec:Data Preparation}) by 0.52 and 0.69 BLEU points (or 0.46 and 0.43 chrF), respectively.
    
    We also propose a novel method for introducing the source graph information during the decoding phase, namely through the cross-attention layer in the decoder (see \S\ref{subsec:Same-Scene-Aware Cross-Attention}). We find that it improves over the baseline and syntactic models, although SASA is generally better. Interestingly, for En-Fi, this model also outperforms SASA, suggesting that some language pairs may benefit more from semantic injection into the decoder.
    
    Overall, through a series of experiments (see~\S\ref{sec:Experiments}), we show the potential of semantics as an aid for NMT. We experiment with a large set of variants of our method, to see where and in what incorporation method they best help. Finally, we show that semantic models outperform UD baselines and can be complementary to them in distant languages, showing improvement when combined. 

\section{Models}\label{sec:Model}

    Transformers have been shown to struggle when translating some types of long-distance dependencies \citep{choshen-abend-2019-automatically, DBLP:journals/corr/abs-2107-06055}, and when facing atypical word order \citep{bisazza2021difficulty}.
    \citet{sulem-etal-2018-semantic} proposed UCCA based preprocessing at inference time, splitting sentences into different scenes. They hypothesized that models need to decompose the input into scenes implicitly, and provide them with such a decomposition, as well as with the original sentence. They show that this may facilitate machine translation \citep{sulem-etal-2020-semantic} and sentence simplification \citep{sulem-etal-2018-simple} in some cases. 
    
    Motivated by these advances, we integrate UCCA to split the source into scenes. 
    However, unlike Sulem et al., we do not alter the sentence length in pre-processing, as this method allows less flexibility in the way information is passed, and as preliminary results in reimplementing this method yielded inferior results (see \S\ref{subsec:SemSplit}). Instead, we investigate ways to integrate the split into the attention architecture. 

    We follow previous work \citep{bugliarello-okazaki-2020-enhancing} in the way we incorporate our semantic information. In their paper, \citet{bugliarello-okazaki-2020-enhancing} introduced syntax in the form of a parent-aware mask, which was applied before the softmax layer in the encoder's self-attention. We mask in a similar method to introduce semantics. However, \textit{parent} in the UCCA framework is an elusive concept, given that nodes may have multiple parents. Hence, we use a different way to express the semantic information in our mask, i.e., we make it \textit{scene-aware}, rather than \textit{parent-aware}.
    
    Following \citet{sulem-etal-2018-simple}, we divide the source sentence into scenes, using the sentence's UCCA parse. We then define our Scene-Aware mask:
    
    \begin{equation}
    M_C[i,j] =  
    \begin{cases}
        1,& \text{if i,j in the same scene}\\
        0,              & \text{otherwise}
    \end{cases}
    \end{equation}

    Intuitively, an attention head masked this way is allowed to attend to other tokens, as long as they share a scene with the current one.\footnote{In case a token belongs to more than one scene, as is the case with the word ``dog'' in Fig.~\ref{fig:ucca_to_mask_example1}, we allow it to attend to tokens of all the scenes it belongs to.}
    Figure~\ref{fig:ucca_to_masks} demonstrates two examples of such masks, accompanied by their UCCA parse graphs and the segmentation into Scenes from which these masks were generated.

    Our base model is the Transformer \citep{vaswani2017attention}, which we enhance by making the attention layers more scene-aware. We force one\footnote{Initial trials with more than one head did not show further benefit for UCCA based models.} of the heads to attend to words in the same scene which we assume are more likely to be related than words from different scenes. As we replace regular self-attention heads with our scene-aware ones, we maintain the same number of heads and layers as in the baseline. 
    
    
    \subsection{Scene-Aware Self-Attention (SASA)}\label{subsec:Scene-Aware Self-Attention}
    \begin{figure}[h!]
    \centering
        \includegraphics[width=5cm]{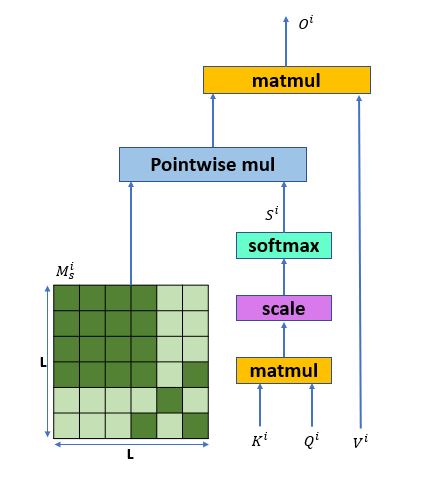}
        \captionsetup{aboveskip=0pt, belowskip=0pt}
        \caption{Scene-aware self-attention head for the input sentence "I saw the dog that barked", consisting of two scenes: "I saw the dog" and "dog barked".}
        \label{fig:model}
    \end{figure}
    
    Figure \ref{fig:model} presents the model's architecture. For a source sentence of length $L$, we obtain the keys, queries, and values matrices denoted by $K^i, Q^i, V^i \in \bb{R}^{L\times d}$, respectively. Then, to get the output matrix $O^i\in \bb{R}^{L\times d}$, we perform the following calculations:
    
        \begin{equation}\label{eq:SASA_eq1}
        S^i = Softmax\left(Q^i\times(K^i)^T\cdot\frac{1}{\sqrt{d_k}}\right)
        \end{equation}
        \begin{equation}\label{eq:SASA_eq2}
        O^i = S^i\odot M_{S}^{i}\times V^i
        \end{equation}
    Where $\frac{1}{\sqrt{d_k}}$ is a scaling factor, the softmax in equation \ref{eq:SASA_eq1} is performed element-wise, $M_{S}^{i}\in {0,1}^{L\times L}$ is our pre-generated scene-aware mask and the $\odot$ in equation \ref{eq:SASA_eq2} denotes an element-wise multiplication. The difference between our method and a vanilla Transformer \citep{vaswani2017attention} lies in equation \ref{eq:SASA_eq2}, with the element-wise multiplication between   $M_{S}^{i}$ and $S^i$, which is absent from the vanilla Transformer (the rest is the same).


\subsection{Scene-Aware Cross-Attention (SACrA)}\label{subsec:Same-Scene-Aware Cross-Attention}

    \begin{figure}[ht]
    \includegraphics[width=7cm]{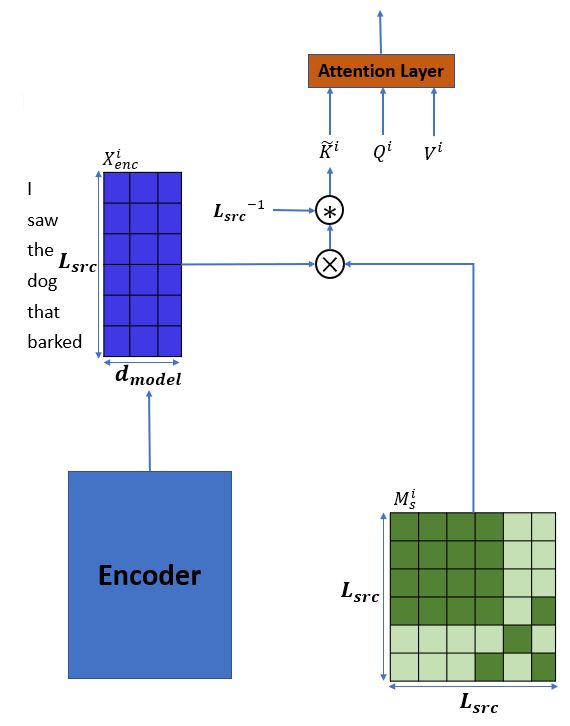}
    \captionsetup{aboveskip=0pt, belowskip=0pt}
    \caption{Scene-aware cross-attention head for the source sentence "I saw the dog that barked."}
    \label{fig:model_cross_attn}
    \end{figure}

    Next, we design a model in which we integrate information about the scene structure through the cross-attention layer in the decoder (see Fig.~\ref{fig:model_cross_attn}). Thus, instead of affecting the overall encoding of the source, we bring forward the splits to aid in selecting the next token. 
    
    Formally, for a source sentence of length $L_{src}$ and target sentence of length $L_{trg}$, we compute for each head the queries and values matrices, denoted by $Q^i\in\bb{R}^{L_{trg}\times d_{model}}$ and $V^i\in\bb{R}^{L_{src}\times d}$, accordingly. Regarding key values, denoted by $\Tilde{K}^i\in\bb{R}^{L_{src}\times L_{trg}}$, we calculate them as follows:
    
        \begin{equation}\label{eq:SACrA_eq}
        \Tilde{K}^i = \left((X^i_{enc})^T\times M_{S}^{i}\right) \cdot \frac{1}{L_{src}}
        \end{equation}
    
    where $X^i_{enc}\in \bb{R}^{L_{src}\times d_{model}}$ is the encoder's output and $M_{S}\in\{0,1\}^{L_{src}\times L_{src}}$ is our pre-generated mask. \\
    Finally, we pass $V^i, Q^i$ and $\Tilde{K}^i$ through a regular attention layer, as with the standard Transformer architecture.


    \paragraph{Scene-Aware Key Matrix.}
    The rationale behind the way we compute our scene-aware keys matrix lies in the role of the keys matrix in an attention layer. In the cross-attention layer, the queries come from the decoder. Source-side contextual information is encoded in the keys, which come from the encoder. Therefore, when we assign the same scene masks to all the words that are included in the same set of scenes, the key values for these words will be the same, and they will thus be treated similarly by the query. As a result, the query will give the same weight to source tokens that share the same set of scenes. Therefore, a complete scene (or a few scenes), rather than specific tokens (as with the vanilla Transformer), will influence what the next generated token will be, which will in turn yield a more scene-aware decoding process.

\section{Experimental Setting}\label{sec:Experimental Settings}

\paragraph{Data Preparation.}\label{sec:Data Preparation}
First, we unescaped HTML characters and tokenized all our parallel corpora \citep{koehn-etal-2007-moses}. 
Next, we removed empty sentences, sentences longer than 100 tokens (either on the source or the target side), sentences with a source-target ratio larger than 1.5, sentences that do not match the corpus's language as determined by langid  \citealp{lui-baldwin-2012-langid}, and sentences that \textit{fast align} \citep{dyer-etal-2013-simple} considers unlikely to align (minimum alignment score of -180). Then, for languages with capitalization, we trained true-casing models on the train set \citep{koehn-etal-2007-moses} and applied them to all inputs to the network. Finally, we trained a BPE model \citep{sennrich-etal-2016-neural}, jointly for language pairs with a similar writing system (e.g., Latin, Cyrillic, etc.)  and separately otherwise,  and then applied them accordingly.

We trained our model on the full WMT16 dataset for the English$\xrightarrow{}$German (En-De) task, using the WMT \textit{newstest2013} as development set. We also trained our models on a train set consisting of Yandex Corpus, News Commentary v15, and Wikititles v2 for the English$\xrightarrow{}$Russian (En-Ru) task. In addition, we trained our models on the full WMT19 dataset (excluding ParaCrawl, in order to avoid noisiness in the data) for the English$\xrightarrow{}$Finnish (En-Fi). Finally, we trained on the full WMT18 dataset for the English$\xrightarrow{}$Turkish (En-Tr) task. For the test sets, we used all the newstests available for every language pair since 2012, excluding the one designated for development.

\paragraph{Models.}
Hyperparameters shared by all models are described in \S\ref{sec:Training Details}. 
We tune the number of heads that we apply the mask to ($\#heads$) and the layers of the encoder we apply SASA to ($layer$), using the En-De development set. We start with tuning the layers for SASA, which we find is $layer=4$, and then we tune the $\#heads$ (while fixing $layer=4$), and get $\#head=1$. We also use the En-De development set to tune the $\#heads$ and the layers of the SACrA model in a similar fashion, namely first the layers and then the $\#heads$ (with the tuned layers fixed).
We find the best hyperparameters are $\#heads=1$ and $layers=2\&3$. For both models, we apply the tuned hyperparameters to all other language pairs. Interestingly, while it is common practice to change all the layers of the model, we find it suboptimal.  
Moreover, the fact that semantic information is more beneficial in higher layers, in contrast to the syntactic information that is most helpful when introduced in lower layers (see \S\ref{sec:Baselines}) may suggest that semantics is relevant for more complex generalization, which is reminiscent of findings by previous work \citep{tenney-etal-2019-bert, Belinkov2018OnIL, tenney2018what, peters-etal-2018-dissecting, blevins-etal-2018-deep, slobodkin-etal-2021-mediators}.

UCCA parses are extracted using a pretrained BERT-based TUPA model, that was trained on sentences in English, German and French \citep{hershcovich2017a}.

\begin{table*}[]
\resizebox{\textwidth}{!}{%
\begin{tabular}{@{}|l|l|l|l|l|l|l|l|l|l|l|@{}}
\toprule
\textbf{models} & \textbf{2012} & \textbf{2013} & \textbf{2014} & \textbf{2015} & \textbf{2016} & \textbf{2017} & \textbf{2018} & \textbf{2019} & \textbf{2020} & \textbf{2020B} \\ \midrule
Transformer & 17.60 & 20.49 & 20.55 & 22.17 & 25.46 & 19.70 & 28.01 & 26.84 & 17.71 & 16.94 \\ \midrule
\begin{tabular}[c]{@{}l@{}}+ binary mask \\ (\#h=1, l= 4)\end{tabular} & \textbf{17.64} & 20.37 & \textbf{20.84} & \textbf{22.48} & 25.32 & 19.76 & \textbf{28.36} & 26.80 & 17.74 & 16.98 \\ \midrule
\begin{tabular}[c]{@{}l@{}}+ scaled mask\\ (\#h=2, l=4, C=0.1)\end{tabular} & 17.41 & 20.21 & 20.53 & 22.43 & 24.95 & \textbf{19.81} & 28.25 & \textbf{27.21} & \textbf{18.03} & \textbf{17.01} \\ \midrule
\begin{tabular}[c]{@{}l@{}}+ normally distributed  mask \\ (\#h=2, l=4, C=$\sqrt{0.5}$)\end{tabular} & 17.39 & \textbf{20.52} & 20.57 & 22.24 & 25.44 & 19.63 & 28.35 & 26.6 & 17.14 & 16.77 \\ \bottomrule
\end{tabular}%
}
\caption{BLEU scores for the top versions of our binary mask, scaled mask, and normally-distributed mask methods across all the WMT En-De newstests.  Each column contains the BLEU scores over the WMT newstest corresponding to the year the column is labeled with (e.g., the scores under column \textit{2015} are for En-De newstest2015). For newstest2020, there was more than one version on WMT, each translated by a different person. Both versions were included, with the second version denoted with a "B". The best score for each test set is boldfaced, unless none is better than the baseline Transformer.}
\label{tab:scaled_masks}
\end{table*}

\paragraph{Binary Mask.}
For the SASA model, we experiment with two types of masks: a binary mask, as described in \S\ref{sec:Model}, and scaled masks, i.e.,

\begin{equation}
M_C[i,j] =  
\begin{cases}
    1,& \text{if i,j in the same scene}\\
    C,              & \text{otherwise}
\end{cases}
\end{equation}

where $C\in (0,1)$.
By doing so, we allow some out-of-scene information to pass through, while still emphasizing the in-scene information (by keeping the value of M for same-scene tokens at 1). In order to tune C, we performed a small grid search over $C\in \{ 0.05, 0.1, 0.15, 0.2, 0.3, 0.5\}$.

Additionally, similarly to  \citet{bugliarello-okazaki-2020-enhancing}, we test a normally-distributed mask, according to the following equation:

\begin{equation}
    M_{i,j}=f_{norm}(x=C\cdot dist(i,j))
\end{equation}

where $f_{norm}$ is the density function of the normal distribution:

    \begin{equation}\label{eq:normal distribution}
        f_{norm}(x)=\frac{1}{\sqrt{2\pi\sigma^2}}e^{-\frac{x^2}{2\sigma^2}}
    \end{equation}

We define a scene-graph where nodes are scenes and edges are drawn between scenes with overlapping words. $dist(i,j)$ is the shortest distance between tokens $i$ and $j$. $\sigma=\frac{1}{\sqrt{2\pi}}$, to ensure the value of $M$ is 1 for words that share a scene ($dist(i,j)$=0), and $C$ is a hyperparameter, which is determined through a grid search over $C\in \{ 0.1, 0.2, 0.5, \sqrt{0.5}\}$. For each of those two scaled versions of the mask, we choose the mask which has the best performance and compare it to the binary mask (see \ref{tab:scaled_masks}). We find that neither outperforms the binary mask. Therefore, we report the rest of our experiments with the binary mask.

\paragraph{Baselines.}\label{sec:Baselines}

We compared our model to a few other models:

\begin{itemize}
    \item \textbf{Transformer.} Standard Transformer-based NMT model, using the standard hyperparameters, as described in \S\ref{sec:Training Details}. 
    \item \textbf{PASCAL.} Following \citet{bugliarello-okazaki-2020-enhancing}, we generate a syntactic mask for the self-attention layer in the encoder. We extract a UD-graph \citep{nivre2016universal}  with udpipe \citep{udpipe:2017}.
    The value of the entries of the masks equal (see equation \ref{eq:normal distribution}): 
    \begin{equation}
        M_{p_t,j}=f_{norm}(x=(j-p_t))
    \end{equation}
    
     with $\sigma=1$ and $p_t$ being the middle position of the $t$-th token's parent in the UD graph of the sentence.
    
    We use the same general hyperparameters as in the Transformer baseline. In addition, following the tuning of \citet{bugliarello-okazaki-2020-enhancing}, we apply the PASCAL mask to five heads of the first attention layer of the encoder, but unlike the original paper, we apply it after the layer's softmax, as it yields better results and also resembles our model's course of action.
    \item \textbf{UDISCAL.} In an attempt to improve the PASCAL model, we generate a mask that instead of only being sensitive to the dependency parent, is sensitive to all the UD relations in the sentences. We denote it UD-Distance-Scaled mask (UDISCAL). Namely, in order to compute the mask, we use a similar equation to that of PASCAL, with a minor alteration:
    \begin{equation}
        M_{i,j}=f_{norm}(x=dist(i,j))
    \end{equation}
    Where $\sigma=1$, and $dist(i,j)$ is defined to be the distance between the token i and the token j in the UD graph of the sentence while treating the graph as undirectional. As with the PASCAL layer, we apply the UD-scaled mask after the softmax layer. But, unlike the PASCAL head, we tuned the architecture's hyperparameters to be just one head of the first layer, after performing a small grid search, namely testing with all layers $l\in[1,4]$, and then with $\#head\in[1,5]$.
\end{itemize}

\paragraph{Training Details.}\label{sec:Training Details}
All our models are based on the standard Transformer-based NMT model \citep{vaswani2017attention}, with 4000 warmup steps. In addition, we use an internal token representation of size 256, per-token cross-entropy loss function, label smoothing with $\epsilon_{l_s}=0.1$ \citep{Szegedy_2016_CVPR}, Adam optimizer, Adam coefficients $\beta_1=0.9$ and $\beta_2=0.98$, and Adam $\epsilon=e^{-1}$. Furthermore, we incorporate 4 layers in the encoder and 4 in the decoder, and we employ a beam-search during inference, with beam size 4 and normalization coefficient $\alpha=0.6$. In addition, we use a batch size of 128 sentences for the training. We use {\it chrF++.py} with 1 word and beta of 3 to obtain chrF+ \citep{Popovic2017chrFWH} score as in WMT19 \citep{ma2019wmt19metrics} and detokenized BLEU \citep{Papineni2002BleuAM} as implemented in Moses. We use the Nematus toolkit \citep{sennrich-EtAl:2017:EACLDemo}, and we train all our models on 4 NVIDIA GPUs for 150K steps.
The average training time for the vanilla Transformer is 21.8 hours, and the average training time for the SASA model is 26.5 hours.


\section{Experiments} \label{sec:Experiments}

\oa{please add copious descriptions to the table captions as to what each column represents, and for Table 1 what language pair it is for}\as{To Omri - is it better now?}\as{to me - If better - change accordingly in the other tables}
We hypothesize that NMT models may benefit from the introduction of semantic structure, and present a set of experiments that support this hypothesis using the above-presented methods.

\subsection{Scene-Aware Self-Attention}\label{subsec:Scene-Aware Self-Attention - Experiments}

\begin{table*}[]
\resizebox{\textwidth}{!}{%
\begin{tabular}{@{}lll@{}}
 & \textbf{Source sentences and Translations} & \textbf{Literal Translations into English} \\ \midrule
\textbf{SRC} & \multicolumn{1}{l|}{I promised a show ?} &  \\ \cmidrule(l){2-3} 
\textbf{BASE} & \multicolumn{1}{l|}{\foreignlanguage{russian}{Я обещал \underline{показать}?}} & I promised \underline{to show}? \\ \cmidrule(l){2-3} 
\textbf{SASA} & \multicolumn{1}{l|}{\foreignlanguage{russian}{Я обещал \underline{шоу}?}} & I promised \underline{a show}? \\ \midrule
\textbf{SRC} & \multicolumn{1}{l|}{Students said they looked forward to his class .} &  \\ \cmidrule(l){2-3} 
\textbf{BASE} & \multicolumn{1}{l|}{\begin{tabular}[c]{@{}l@{}}\foreignlanguage{russian}{Студенты сказали, что они} \\ \foreignlanguage{russian}{\underline{смотрят на} свой класс.}\end{tabular}} & \begin{tabular}[c]{@{}l@{}}Students said, that they \\ \underline{look at} one's classroom.\end{tabular} \\ \cmidrule(l){2-3} 
\textbf{SASA} & \multicolumn{1}{l|}{\begin{tabular}[c]{@{}l@{}}\foreignlanguage{russian}{Студенты сказали, что они} \\ \foreignlanguage{russian}{\underline{с нетерпением ждали} своего класса.}\end{tabular}} & \begin{tabular}[c]{@{}l@{}}Students said, that they \\ \underline{impatiently waited} one's classroom.\end{tabular} \\ \midrule
\textbf{SRC} & \multicolumn{1}{l|}{\begin{tabular}[c]{@{}l@{}}I remember those kids I used to play \\ with in the yard who never got out .\end{tabular}} &  \\ \cmidrule(l){2-3} 
\textbf{BASE} & \multicolumn{1}{l|}{\begin{tabular}[c]{@{}l@{}}\foreignlanguage{russian}{Я помню тех детей, которые я играл} \\ \foreignlanguage{russian}{\underline{с двором},  \underline{который} никогда не \underline{выходил}.}\end{tabular}} & \begin{tabular}[c]{@{}l@{}}I remember those kids, that I played \underline{with yard}, \underline{that}\\ never \underline{got out} ("that" and "got out" refer to yard).\end{tabular} \\ \cmidrule(l){2-3} 
\textbf{SASA} & \multicolumn{1}{l|}{\begin{tabular}[c]{@{}l@{}}\foreignlanguage{russian}{Я помню тех детей, с которыми я играл} \\ \foreignlanguage{russian}{\underline{на дворе}, \underline{которые} никогда не \underline{вышли}.}\end{tabular}} & \begin{tabular}[c]{@{}l@{}}I remember those kids, with which I played \underline{in yard}, \\ \underline{that} never \underline{got out} ("that" and "got out" refer to kids).\end{tabular} \\ \bottomrule
\end{tabular}%
}
\caption{Examples of correct translations generated by SASA, compared to the baseline Transformer.}
\label{tab:SASA_vs_base examples}
\end{table*}

We find that on average, SASA outperforms the Transformer for all four language pairs (see \ref{tab:full_testsets_Bleu}), at times having gains larger than 1 BLEU point. Moreover, we assess the consistency of SASA's gains, using the sign-test, and get a p-value smaller than 0.01, thus exhibiting a statistically significant improvement (see \S\ref{subsec:sign_test_full}). We see a similar trend when evaluating the performance using the chrF metric (see \S\ref{subsec:ChrF_Results}), which further highlights our model's consistent gains.

We also evaluate our model's performance on sentences with long dependencies (see \ref{subsec:Challenge_Sets}), which were found to pose a challenge for Transformers \citep{choshen-abend-2019-automatically}. We assume that such cases could benefit greatly from the semantic introduction. In contrast to our hypothesis, we find the gain to be only slightly larger than in the general case, which leads us to conclude the improvements we see do not specifically originate from the syntactic challenge. Nevertheless, we still observe a consistent improvement, with gains of up to 1.41 BLEU points, which further underscores our model's superiority over the baseline model.

\paragraph{Qualitative Analysis.}
Table \ref{tab:SASA_vs_base examples} presents a few examples in which the baseline Transformer errs, whereas our model translates correctly (see \S\ref{subsec:Qualitative Analysis - UCCA Parsings} for the UCCA parsings of the examples). In the first example, the Transformer translates the word ``show'' as a verb, i.e. \textit{to show}, rather than as a noun. In the second example, the baseline model makes two errors: it misinterprets the word "look forward to" as "look at", and it also translates it as a present-tense verb rather than past-tense. The third example is particularly interesting, as it highlights our model's strength. In this example, the Transformer makes two mistakes: first, it translates the part "play with (someone) in the yard"
as "play with the yard". Next, it attributes the descriptive clause "which never got out" to the yard, rather than the children. It seems then that introducing information about the \textit{scene} structure into the model facilitates the translation, since it both groups the word "kids" with the phrase "I used to play with in the yard", and it also separates "never got out" from the word "yard". Instead, it clusters the latter with "kids", thus highlighting the relations between words in the sentence. In general, all these examples are cases where the network succeeds in disambiguating a word in its context.

\subsection{Comparison to Syntactic Masks}\label{subsec:Comparison to Syntactic Masks}

\begin{table*}[]
\resizebox{\textwidth}{!}{%
\begin{tabular}{@{}lccccccccclc@{}}
\rowcolor[HTML]{FFFFFF} 
\multicolumn{11}{c}{\cellcolor[HTML]{FFFFFF}{\color[HTML]{000000} {\ul \textbf{En-De}}}} & {\color[HTML]{000000} } \\ \cmidrule(r){1-11}
\multicolumn{1}{|l|}{{\color[HTML]{000000} \textbf{models}}} & \multicolumn{1}{c|}{{\color[HTML]{000000} \textbf{2012}}} & \multicolumn{1}{c|}{\textbf{2014}} & \multicolumn{1}{c|}{{\color[HTML]{000000} \textbf{2015}}} & \multicolumn{1}{c|}{{\color[HTML]{000000} \textbf{2016}}} & \multicolumn{1}{c|}{{\color[HTML]{000000} \textbf{2017}}} & \multicolumn{1}{c|}{{\color[HTML]{000000} \textbf{2018}}} & \multicolumn{1}{c|}{{\color[HTML]{000000} \textbf{2019}}} & \multicolumn{1}{c|}{{\color[HTML]{000000} \textbf{2020}}} & \multicolumn{1}{c|}{{\color[HTML]{000000} \textbf{2020B}}} & \multicolumn{1}{c|}{{\color[HTML]{000000} {\ul \textbf{average}}}} & {\color[HTML]{000000} {\ul \textbf{}}} \\ \cmidrule(r){1-11}
\multicolumn{1}{|l|}{{\color[HTML]{000000} Transformer}} & \multicolumn{1}{c|}{{\color[HTML]{000000} 17.6}} & \multicolumn{1}{c|}{20.55} & \multicolumn{1}{c|}{{\color[HTML]{000000} 22.17}} & \multicolumn{1}{c|}{{\color[HTML]{000000} \textbf{25.46}}} & \multicolumn{1}{c|}{{\color[HTML]{000000} 19.7}} & \multicolumn{1}{c|}{{\color[HTML]{000000} 28.01}} & \multicolumn{1}{c|}{{\color[HTML]{000000} 26.84}} & \multicolumn{1}{c|}{{\color[HTML]{000000} 17.71}} & \multicolumn{1}{c|}{{\color[HTML]{000000} 16.94}} & \multicolumn{1}{c|}{{\color[HTML]{000000} 21.66}} & {\color[HTML]{000000} } \\ \cmidrule(r){1-11}
\multicolumn{1}{|l|}{{\color[HTML]{000000} PASCAL}} & \multicolumn{1}{c|}{{\color[HTML]{000000} 17.34}} & \multicolumn{1}{c|}{20.59} & \multicolumn{1}{c|}{{\color[HTML]{000000} \textbf{22.62}}} & \multicolumn{1}{c|}{{\color[HTML]{000000} 25.1}} & \multicolumn{1}{c|}{{\color[HTML]{000000} 19.92}} & \multicolumn{1}{c|}{{\color[HTML]{000000} 28.09}} & \multicolumn{1}{c|}{{\color[HTML]{000000} 26.61}} & \multicolumn{1}{c|}{{\color[HTML]{000000} 17.5}} & \multicolumn{1}{c|}{{\color[HTML]{000000} 16.81}} & \multicolumn{1}{c|}{{\color[HTML]{000000} 21.62}} & {\color[HTML]{000000} } \\ \cmidrule(r){1-11}
\multicolumn{1}{|l|}{{\color[HTML]{000000} UDISCAL}} & \multicolumn{1}{c|}{{\color[HTML]{000000} 17.42}} & \multicolumn{1}{c|}{20.86} & \multicolumn{1}{c|}{{\color[HTML]{000000} 22.53}} & \multicolumn{1}{c|}{{\color[HTML]{000000} 25.23}} & \multicolumn{1}{c|}{{\color[HTML]{000000} \textbf{19.95}}} & \multicolumn{1}{c|}{{\color[HTML]{000000} 27.87}} & \multicolumn{1}{c|}{{\color[HTML]{000000} 26.8}} & \multicolumn{1}{c|}{{\color[HTML]{000000} 17.06}} & \multicolumn{1}{c|}{{\color[HTML]{000000} 16.39}} & \multicolumn{1}{c|}{{\color[HTML]{000000} 21.57}} & {\color[HTML]{000000} } \\ \cmidrule(r){1-11}
\multicolumn{1}{|l|}{{\color[HTML]{000000} SASA}} & \multicolumn{1}{c|}{{\color[HTML]{000000} \textbf{17.64$^\uparrow$}}} & \multicolumn{1}{c|}{20.84} & \multicolumn{1}{c|}{{\color[HTML]{000000} 22.48}} & \multicolumn{1}{c|}{{\color[HTML]{000000} 25.32}} & \multicolumn{1}{c|}{{\color[HTML]{000000} 19.76}} & \multicolumn{1}{c|}{{\color[HTML]{000000} \textbf{28.36$^\uparrow$}}} & \multicolumn{1}{c|}{{\color[HTML]{000000} 26.8}} & \multicolumn{1}{c|}{{\color[HTML]{000000} \textbf{17.74$^\uparrow$}}} & \multicolumn{1}{c|}{{\color[HTML]{000000} \textbf{16.98$^\uparrow$}}} & \multicolumn{1}{c|}{{\color[HTML]{000000} \textbf{21.77$^\uparrow$}}} & {\color[HTML]{000000} \textbf{}} \\ \cmidrule(r){1-11}
\multicolumn{1}{|l|}{{\color[HTML]{000000} SASA + UDISCAL}} & \multicolumn{1}{c|}{{\color[HTML]{000000} 17.51}} & \multicolumn{1}{c|}{20.42} & \multicolumn{1}{c|}{{\color[HTML]{000000} 22.1}} & \multicolumn{1}{c|}{{\color[HTML]{000000} 24.9}} & \multicolumn{1}{c|}{{\color[HTML]{000000} 19.72}} & \multicolumn{1}{c|}{{\color[HTML]{000000} 28.35}} & \multicolumn{1}{c|}{{\color[HTML]{000000} \textbf{27.14$^{*}$}}} & \multicolumn{1}{c|}{{\color[HTML]{000000} 17.59}} & \multicolumn{1}{c|}{{\color[HTML]{000000} 16.68}} & \multicolumn{1}{c|}{{\color[HTML]{000000} 21.60}} & {\color[HTML]{000000} } \\ \cmidrule(r){1-11}
\multicolumn{1}{|l|}{{\color[HTML]{000000} SACrA}} & \multicolumn{1}{c|}{{\color[HTML]{000000} 17.11}} & \multicolumn{1}{c|}{20.9$^\uparrow$} & \multicolumn{1}{c|}{{\color[HTML]{000000} 22.59}} & \multicolumn{1}{c|}{{\color[HTML]{000000} 24.64}} & \multicolumn{1}{c|}{{\color[HTML]{000000} 19.79}} & \multicolumn{1}{c|}{{\color[HTML]{000000} 27.88}} & \multicolumn{1}{c|}{{\color[HTML]{000000} 26.28}} & \multicolumn{1}{c|}{{\color[HTML]{000000} 16.8}} & \multicolumn{1}{c|}{{\color[HTML]{000000} 16.25}} & \multicolumn{1}{c|}{{\color[HTML]{000000} 21.36}} & {\color[HTML]{000000} } \\ \cmidrule(r){1-11}
\multicolumn{1}{|l|}{{\color[HTML]{000000} SACrA + UDISCAL}} & \multicolumn{1}{c|}{{\color[HTML]{000000} 17.07}} & \multicolumn{1}{c|}{\textbf{21.09$^{*}$}} & \multicolumn{1}{c|}{{\color[HTML]{000000} 22.26}} & \multicolumn{1}{c|}{{\color[HTML]{000000} 24.85}} & \multicolumn{1}{c|}{{\color[HTML]{000000} 19.56}} & \multicolumn{1}{c|}{{\color[HTML]{000000} 28.1$^{*}$}} & \multicolumn{1}{c|}{{\color[HTML]{000000} 26.49}} & \multicolumn{1}{c|}{{\color[HTML]{000000} 16.66}} & \multicolumn{1}{c|}{{\color[HTML]{000000} 15.93}} & \multicolumn{1}{c|}{{\color[HTML]{000000} 21.33}} & {\color[HTML]{000000} } \\ \cmidrule(r){1-11}
\rowcolor[HTML]{FFFFFF} 
\multicolumn{12}{c}{\cellcolor[HTML]{FFFFFF}{\color[HTML]{000000} {\ul \textbf{En-Ru}}}} \\ \midrule
\multicolumn{1}{|l|}{{\color[HTML]{000000} \textbf{models}}} & \multicolumn{1}{c|}{{\color[HTML]{000000} \textbf{2012}}} & \multicolumn{1}{c|}{\textbf{2013}} & \multicolumn{1}{c|}{{\color[HTML]{000000} \textbf{2014}}} & \multicolumn{1}{c|}{{\color[HTML]{000000} \textbf{2015}}} & \multicolumn{1}{c|}{{\color[HTML]{000000} \textbf{2016}}} & \multicolumn{1}{c|}{{\color[HTML]{000000} \textbf{2017}}} & \multicolumn{1}{c|}{{\color[HTML]{000000} \textbf{2018}}} & \multicolumn{1}{c|}{{\color[HTML]{000000} \textbf{2019}}} & \multicolumn{1}{c|}{{\color[HTML]{000000} \textbf{2020}}} & \multicolumn{1}{c|}{{\color[HTML]{000000} \textbf{2020B}}} & \multicolumn{1}{c|}{{\color[HTML]{000000} {\ul \textbf{average}}}} \\ \midrule
\multicolumn{1}{|l|}{{\color[HTML]{000000} Transformer}} & \multicolumn{1}{c|}{{\color[HTML]{000000} 24.32}} & \multicolumn{1}{c|}{{\color[HTML]{000000} 18.11}} & \multicolumn{1}{c|}{{\color[HTML]{000000} 25.35}} & \multicolumn{1}{c|}{{\color[HTML]{000000} 21.1}} & \multicolumn{1}{c|}{{\color[HTML]{000000} 19.77}} & \multicolumn{1}{c|}{{\color[HTML]{000000} 22.34}} & \multicolumn{1}{c|}{{\color[HTML]{000000} 19}} & \multicolumn{1}{c|}{{\color[HTML]{000000} 20.14}} & \multicolumn{1}{c|}{{\color[HTML]{000000} 15.64}} & \multicolumn{1}{c|}{{\color[HTML]{000000} 22.33}} & \multicolumn{1}{c|}{{\color[HTML]{000000} 20.81}} \\ \midrule
\multicolumn{1}{|l|}{{\color[HTML]{000000} PASCAL}} & \multicolumn{1}{c|}{{\color[HTML]{000000} 23.78}} & \multicolumn{1}{c|}{{\color[HTML]{000000} 18.37}} & \multicolumn{1}{c|}{{\color[HTML]{000000} 24.87}} & \multicolumn{1}{c|}{{\color[HTML]{000000} 20.97}} & \multicolumn{1}{c|}{{\color[HTML]{000000} 19.81}} & \multicolumn{1}{c|}{{\color[HTML]{000000} 21.83}} & \multicolumn{1}{c|}{{\color[HTML]{000000} 18.81}} & \multicolumn{1}{c|}{{\color[HTML]{000000} 19.93}} & \multicolumn{1}{c|}{{\color[HTML]{000000} 15.42}} & \multicolumn{1}{c|}{{\color[HTML]{000000} 21.48}} & \multicolumn{1}{c|}{{\color[HTML]{000000} 20.53}} \\ \midrule
\multicolumn{1}{|l|}{{\color[HTML]{000000} UDISCAL}} & \multicolumn{1}{c|}{{\color[HTML]{000000} 23.88}} & \multicolumn{1}{c|}{{\color[HTML]{000000} 18.31}} & \multicolumn{1}{c|}{{\color[HTML]{000000} 25.23}} & \multicolumn{1}{c|}{{\color[HTML]{000000} 20.82}} & \multicolumn{1}{c|}{{\color[HTML]{000000} \textbf{20.31}}} & \multicolumn{1}{c|}{{\color[HTML]{000000} 22.15}} & \multicolumn{1}{c|}{{\color[HTML]{000000} 19.27}} & \multicolumn{1}{c|}{{\color[HTML]{000000} 20.32}} & \multicolumn{1}{c|}{{\color[HTML]{000000} 15.7}} & \multicolumn{1}{c|}{{\color[HTML]{000000} 22.19}} & \multicolumn{1}{c|}{{\color[HTML]{000000} 20.82}} \\ \midrule
\multicolumn{1}{|l|}{{\color[HTML]{000000} SASA}} & \multicolumn{1}{c|}{{\color[HTML]{000000} 24.17}} & \multicolumn{1}{c|}{{\color[HTML]{000000} \textbf{18.43$^\uparrow$}}} & \multicolumn{1}{c|}{{\color[HTML]{000000} \textbf{25.53$^\uparrow$}}} & \multicolumn{1}{c|}{{\color[HTML]{000000} \textbf{21.59$^\uparrow$}}} & \multicolumn{1}{c|}{{\color[HTML]{000000} 20.11}} & \multicolumn{1}{c|}{{\color[HTML]{000000} \textbf{22.69$^\uparrow$}}} & \multicolumn{1}{c|}{{\color[HTML]{000000} \textbf{19.53$^\uparrow$}}} & \multicolumn{1}{c|}{{\color[HTML]{000000} 20.2}} & \multicolumn{1}{c|}{{\color[HTML]{000000} 15.76$^\uparrow$}} & \multicolumn{1}{c|}{{\color[HTML]{000000} \textbf{23.36$^\uparrow$}}} & \multicolumn{1}{c|}{{\color[HTML]{000000} \textbf{21.14$^\uparrow$}}} \\ \midrule
\multicolumn{1}{|l|}{\cellcolor[HTML]{FFFFFF}{\color[HTML]{000000} SASA + UDISCAL}} & \multicolumn{1}{c|}{\cellcolor[HTML]{FFFFFF}{\color[HTML]{000000} \textbf{24.36$^{*}$}}} & \multicolumn{1}{c|}{{\color[HTML]{000000} 18.29}} & \multicolumn{1}{c|}{\cellcolor[HTML]{FFFFFF}{\color[HTML]{000000} 25.43}} & \multicolumn{1}{c|}{\cellcolor[HTML]{FFFFFF}{\color[HTML]{000000} 21.01}} & \multicolumn{1}{c|}{\cellcolor[HTML]{FFFFFF}{\color[HTML]{000000} 19.79}} & \multicolumn{1}{c|}{\cellcolor[HTML]{FFFFFF}{\color[HTML]{000000} 22.49}} & \multicolumn{1}{c|}{\cellcolor[HTML]{FFFFFF}{\color[HTML]{000000} 19.25}} & \multicolumn{1}{c|}{\cellcolor[HTML]{FFFFFF}{\color[HTML]{000000} \textbf{20.4$^{*}$}}} & \multicolumn{1}{c|}{\cellcolor[HTML]{FFFFFF}{\color[HTML]{000000} \textbf{15.97$^{*}$}}} & \multicolumn{1}{c|}{\cellcolor[HTML]{FFFFFF}{\color[HTML]{000000} 22.42}} & \multicolumn{1}{c|}{{\color[HTML]{000000} 20.94}} \\ \midrule
\multicolumn{1}{|l|}{{\color[HTML]{000000} SACrA}} & \multicolumn{1}{c|}{{\color[HTML]{000000} 24.12}} & \multicolumn{1}{c|}{{\color[HTML]{000000} 18.24}} & \multicolumn{1}{c|}{{\color[HTML]{000000} 25.43$^\uparrow$}} & \multicolumn{1}{c|}{{\color[HTML]{000000} 21}} & \multicolumn{1}{c|}{{\color[HTML]{000000} 20.07}} & \multicolumn{1}{c|}{{\color[HTML]{000000} 22.49$^\uparrow$}} & \multicolumn{1}{c|}{{\color[HTML]{000000} 19.3$^\uparrow$}} & \multicolumn{1}{c|}{{\color[HTML]{000000} 20.18}} & \multicolumn{1}{c|}{{\color[HTML]{000000} 15.79$^\uparrow$}} & \multicolumn{1}{c|}{{\color[HTML]{000000} 22.15}} & \multicolumn{1}{c|}{{\color[HTML]{000000} 20.88$^\uparrow$}} \\ \midrule
\multicolumn{1}{|l|}{{\color[HTML]{000000} SACrA + UDISCAL}} & \multicolumn{1}{c|}{{\color[HTML]{000000} 23.54}} & \multicolumn{1}{c|}{{\color[HTML]{000000} 17.99}} & \multicolumn{1}{c|}{{\color[HTML]{000000} 24.91}} & \multicolumn{1}{c|}{{\color[HTML]{000000} 20.62}} & \multicolumn{1}{c|}{{\color[HTML]{000000} 19.67}} & \multicolumn{1}{c|}{{\color[HTML]{000000} 21.55}} & \multicolumn{1}{c|}{{\color[HTML]{000000} 18.63}} & \multicolumn{1}{c|}{{\color[HTML]{000000} 19.89}} & \multicolumn{1}{c|}{{\color[HTML]{000000} 15.64}} & \multicolumn{1}{c|}{{\color[HTML]{000000} 20.79}} & \multicolumn{1}{c|}{{\color[HTML]{000000} 20.32}} \\ \midrule
\multicolumn{9}{c}{\cellcolor[HTML]{FFFFFF}{\color[HTML]{000000} {\ul \textbf{En-Fi}}}} & \cellcolor[HTML]{FFFFFF}{\color[HTML]{000000} } & {\color[HTML]{000000} } & \multicolumn{1}{l}{{\color[HTML]{000000} }} \\ \cmidrule(r){1-9}
\multicolumn{1}{|l|}{{\color[HTML]{000000} \textbf{models}}} & \multicolumn{1}{c|}{{\color[HTML]{000000} \textbf{2015}}} & \multicolumn{1}{c|}{\textbf{2016}} & \multicolumn{1}{c|}{{\color[HTML]{000000} \textbf{2016B}}} & \multicolumn{1}{c|}{{\color[HTML]{000000} \textbf{2017}}} & \multicolumn{1}{c|}{{\color[HTML]{000000} \textbf{2017B}}} & \multicolumn{1}{c|}{{\color[HTML]{000000} \textbf{2018}}} & \multicolumn{1}{c|}{{\color[HTML]{000000} \textbf{2019}}} & \multicolumn{1}{c|}{{\color[HTML]{000000} {\ul \textbf{average}}}} & {\color[HTML]{000000} {\ul \textbf{}}} & {\color[HTML]{000000} \textbf{}} & \multicolumn{1}{l}{{\color[HTML]{000000} \textbf{}}} \\ \cmidrule(r){1-9}
\multicolumn{1}{|l|}{{\color[HTML]{000000} Transformer}} & \multicolumn{1}{c|}{{\color[HTML]{000000} 11.22}} & \multicolumn{1}{c|}{12.76} & \multicolumn{1}{c|}{{\color[HTML]{000000} 10.2}} & \multicolumn{1}{c|}{{\color[HTML]{000000} 13.35}} & \multicolumn{1}{c|}{{\color[HTML]{000000} 11.37}} & \multicolumn{1}{c|}{{\color[HTML]{000000} 9.32}} & \multicolumn{1}{c|}{{\color[HTML]{000000} 12.21}} & \multicolumn{1}{c|}{{\color[HTML]{000000} 11.49}} & {\color[HTML]{000000} } & {\color[HTML]{000000} } & \multicolumn{1}{l}{{\color[HTML]{000000} }} \\ \cmidrule(r){1-9}
\multicolumn{1}{|l|}{{\color[HTML]{000000} PASCAL}} & \multicolumn{1}{c|}{{\color[HTML]{000000} 11.2}} & \multicolumn{1}{c|}{12.67} & \multicolumn{1}{c|}{{\color[HTML]{000000} 10.13}} & \multicolumn{1}{c|}{{\color[HTML]{000000} 13.54}} & \multicolumn{1}{c|}{{\color[HTML]{000000} 11.24}} & \multicolumn{1}{c|}{{\color[HTML]{000000} 9.62}} & \multicolumn{1}{c|}{{\color[HTML]{000000} 12.23}} & \multicolumn{1}{c|}{{\color[HTML]{000000} 11.52}} & {\color[HTML]{000000} \textbf{}} & {\color[HTML]{000000} } & \multicolumn{1}{l}{{\color[HTML]{000000} }} \\ \cmidrule(r){1-9}
\multicolumn{1}{|l|}{{\color[HTML]{000000} UDISCAL}} & \multicolumn{1}{c|}{{\color[HTML]{000000} 10.87}} & \multicolumn{1}{c|}{12.78} & \multicolumn{1}{c|}{{\color[HTML]{000000} 10.23}} & \multicolumn{1}{c|}{{\color[HTML]{000000} 13.51}} & \multicolumn{1}{c|}{{\color[HTML]{000000} 11.43}} & \multicolumn{1}{c|}{{\color[HTML]{000000} 9.2}} & \multicolumn{1}{c|}{{\color[HTML]{000000} 11.99}} & \multicolumn{1}{c|}{{\color[HTML]{000000} 11.43}} & {\color[HTML]{000000} } & {\color[HTML]{000000} \textbf{}} & \multicolumn{1}{l}{{\color[HTML]{000000} \textbf{}}} \\ \cmidrule(r){1-9}
\multicolumn{1}{|l|}{{\color[HTML]{000000} SASA}} & \multicolumn{1}{c|}{{\color[HTML]{000000} 11.37$^\uparrow$}} & \multicolumn{1}{c|}{\textbf{12.88$^\uparrow$}} & \multicolumn{1}{c|}{{\color[HTML]{000000} \textbf{10.52$^\uparrow$}}} & \multicolumn{1}{c|}{{\color[HTML]{000000} 13.74$^\uparrow$}} & \multicolumn{1}{c|}{{\color[HTML]{000000} 11.5$^\uparrow$}} & \multicolumn{1}{c|}{{\color[HTML]{000000} 9.56}} & \multicolumn{1}{c|}{{\color[HTML]{000000} 12.12}} & \multicolumn{1}{c|}{{\color[HTML]{000000} 11.67$^\uparrow$}} & {\color[HTML]{000000} \textbf{}} & {\color[HTML]{000000} \textbf{}} & \multicolumn{1}{l}{{\color[HTML]{000000} \textbf{}}} \\ \cmidrule(r){1-9}
\multicolumn{1}{|l|}{{\color[HTML]{000000} SASA + UDISCAL}} & \multicolumn{1}{c|}{\cellcolor[HTML]{FFFFFF}{\color[HTML]{000000} \textbf{11.56$^{*}$}}} & \multicolumn{1}{c|}{\cellcolor[HTML]{FFFFFF}12.8} & \multicolumn{1}{c|}{\cellcolor[HTML]{FFFFFF}{\color[HTML]{000000} 10.28}} & \multicolumn{1}{c|}{\cellcolor[HTML]{FFFFFF}{\color[HTML]{000000} \textbf{13.91$^{*}$}}} & \multicolumn{1}{c|}{\cellcolor[HTML]{FFFFFF}{\color[HTML]{000000} \textbf{11.52$^{*}$}}} & \multicolumn{1}{c|}{\cellcolor[HTML]{FFFFFF}{\color[HTML]{000000} \textbf{9.75$^{*}$}}} & \multicolumn{1}{c|}{\cellcolor[HTML]{FFFFFF}{\color[HTML]{000000} \textbf{12.64$^{*}$}}} & \multicolumn{1}{c|}{\cellcolor[HTML]{FFFFFF}{\color[HTML]{000000} \textbf{11.78$^{*}$}}} & {\color[HTML]{000000} \textbf{}} & \cellcolor[HTML]{FFFFFF}{\color[HTML]{000000} \textbf{}} & \multicolumn{1}{l}{\cellcolor[HTML]{FFFFFF}{\color[HTML]{000000} \textbf{}}} \\ \cmidrule(r){1-9}
\multicolumn{1}{|l|}{{\color[HTML]{000000} SACrA}} & \multicolumn{1}{c|}{{\color[HTML]{000000} 11.48$^\uparrow$}} & \multicolumn{1}{c|}{12.86$^\uparrow$} & \multicolumn{1}{c|}{{\color[HTML]{000000} 10.41$^\uparrow$}} & \multicolumn{1}{c|}{{\color[HTML]{000000} 13.66$^\uparrow$}} & \multicolumn{1}{c|}{{\color[HTML]{000000} 11.49$^\uparrow$}} & \multicolumn{1}{c|}{{\color[HTML]{000000} 9.62}} & \multicolumn{1}{c|}{{\color[HTML]{000000} 12.51$^\uparrow$}} & \multicolumn{1}{c|}{{\color[HTML]{000000} 11.72$^\uparrow$}} & {\color[HTML]{000000} \textbf{}} & {\color[HTML]{000000} \textbf{}} & \multicolumn{1}{l}{{\color[HTML]{000000} \textbf{}}} \\ \cmidrule(r){1-9}
\multicolumn{1}{|l|}{{\color[HTML]{000000} SACrA + UDISCAL}} & \multicolumn{1}{c|}{{\color[HTML]{000000} 11.06}} & \multicolumn{1}{c|}{12.6} & \multicolumn{1}{c|}{{\color[HTML]{000000} 10.13}} & \multicolumn{1}{c|}{{\color[HTML]{000000} 13.43}} & \multicolumn{1}{c|}{{\color[HTML]{000000} 11.26}} & \multicolumn{1}{c|}{{\color[HTML]{000000} 9.23}} & \multicolumn{1}{c|}{{\color[HTML]{000000} 12.05}} & \multicolumn{1}{c|}{{\color[HTML]{000000} 11.39}} & {\color[HTML]{000000} } & {\color[HTML]{000000} } & \multicolumn{1}{l}{{\color[HTML]{000000} }} \\ \cmidrule(r){1-9}
\multicolumn{5}{c}{\cellcolor[HTML]{FFFFFF}{\color[HTML]{000000} {\ul \textbf{En-Tr}}}} & \cellcolor[HTML]{FFFFFF}{\color[HTML]{000000} } & {\color[HTML]{000000} } & {\color[HTML]{000000} } & {\color[HTML]{000000} } & {\color[HTML]{000000} } & \multicolumn{1}{c}{{\color[HTML]{000000} }} & {\color[HTML]{000000} } \\ \cmidrule(r){1-5}
\multicolumn{1}{|l|}{{\color[HTML]{000000} \textbf{models}}} & \multicolumn{1}{c|}{{\color[HTML]{000000} \textbf{2016}}} & \multicolumn{1}{c|}{\textbf{2017}} & \multicolumn{1}{c|}{{\color[HTML]{000000} \textbf{2018}}} & \multicolumn{1}{c|}{{\color[HTML]{000000} {\ul \textbf{average}}}} & {\color[HTML]{000000} {\ul \textbf{}}} & \multicolumn{1}{l}{{\color[HTML]{000000} }} & \multicolumn{1}{l}{{\color[HTML]{000000} }} & \multicolumn{1}{l}{{\color[HTML]{000000} }} & \multicolumn{1}{l}{{\color[HTML]{000000} }} & {\color[HTML]{000000} } & \multicolumn{1}{l}{{\color[HTML]{000000} }} \\ \cmidrule(r){1-5}
\multicolumn{1}{|l|}{{\color[HTML]{000000} Transformer}} & \multicolumn{1}{c|}{{\color[HTML]{000000} 8.43}} & \multicolumn{1}{c|}{8.55} & \multicolumn{1}{c|}{{\color[HTML]{000000} 8.1}} & \multicolumn{1}{c|}{{\color[HTML]{000000} 8.36}} & {\color[HTML]{000000} } & \multicolumn{1}{l}{{\color[HTML]{000000} }} & \multicolumn{1}{l}{{\color[HTML]{000000} }} & \multicolumn{1}{l}{{\color[HTML]{000000} }} & \multicolumn{1}{l}{{\color[HTML]{000000} }} & {\color[HTML]{000000} } & \multicolumn{1}{l}{{\color[HTML]{000000} }} \\ \cmidrule(r){1-5}
\multicolumn{1}{|l|}{{\color[HTML]{000000} PASCAL}} & \multicolumn{1}{c|}{{\color[HTML]{000000} 8.5}} & \multicolumn{1}{c|}{8.76} & \multicolumn{1}{c|}{{\color[HTML]{000000} 7.98}} & \multicolumn{1}{c|}{{\color[HTML]{000000} 8.41}} & {\color[HTML]{000000} \textbf{}} & \multicolumn{1}{l}{{\color[HTML]{000000} }} & \multicolumn{1}{l}{{\color[HTML]{000000} }} & \multicolumn{1}{l}{{\color[HTML]{000000} }} & \multicolumn{1}{l}{{\color[HTML]{000000} }} & {\color[HTML]{000000} } & \multicolumn{1}{l}{{\color[HTML]{000000} }} \\ \cmidrule(r){1-5}
\multicolumn{1}{|l|}{{\color[HTML]{000000} UDISCAL}} & \multicolumn{1}{c|}{{\color[HTML]{000000} 8.33}} & \multicolumn{1}{c|}{8.66} & \multicolumn{1}{c|}{{\color[HTML]{000000} 8.03}} & \multicolumn{1}{c|}{{\color[HTML]{000000} 8.34}} & {\color[HTML]{000000} } & \multicolumn{1}{l}{{\color[HTML]{000000} }} & \multicolumn{1}{l}{{\color[HTML]{000000} }} & \multicolumn{1}{l}{{\color[HTML]{000000} }} & \multicolumn{1}{l}{{\color[HTML]{000000} }} & {\color[HTML]{000000} } & \multicolumn{1}{l}{{\color[HTML]{000000} }} \\ \cmidrule(r){1-5}
\multicolumn{1}{|l|}{{\color[HTML]{000000} SASA}} & \multicolumn{1}{c|}{{\color[HTML]{000000} 8.59$^\uparrow$}} & \multicolumn{1}{c|}{8.86$^\uparrow$} & \multicolumn{1}{c|}{{\color[HTML]{000000} 8.16$^\uparrow$}} & \multicolumn{1}{c|}{{\color[HTML]{000000} 8.54$^\uparrow$}} & {\color[HTML]{000000} \textbf{}} & \multicolumn{1}{l}{{\color[HTML]{000000} }} & \multicolumn{1}{l}{{\color[HTML]{000000} }} & \multicolumn{1}{l}{{\color[HTML]{000000} }} & \multicolumn{1}{l}{{\color[HTML]{000000} }} & {\color[HTML]{000000} } & \multicolumn{1}{l}{{\color[HTML]{000000} }} \\ \cmidrule(r){1-5}
\multicolumn{1}{|l|}{{\color[HTML]{000000} SASA + UDISCAL}} & \multicolumn{1}{c|}{\cellcolor[HTML]{FFFFFF}{\color[HTML]{000000} \textbf{8.64$^{*}$}}} & \multicolumn{1}{c|}{\cellcolor[HTML]{FFFFFF}\textbf{8.87$^{*}$}} & \multicolumn{1}{c|}{\cellcolor[HTML]{FFFFFF}{\color[HTML]{000000} \textbf{8.2$^{*}$}}} & \multicolumn{1}{c|}{\cellcolor[HTML]{FFFFFF}{\color[HTML]{000000} \textbf{8.57$^{*}$}}} & {\color[HTML]{000000} \textbf{}} & \multicolumn{1}{l}{{\color[HTML]{000000} }} & \multicolumn{1}{l}{{\color[HTML]{000000} }} & \multicolumn{1}{l}{{\color[HTML]{000000} }} & \multicolumn{1}{l}{{\color[HTML]{000000} }} & {\color[HTML]{000000} } & \multicolumn{1}{l}{{\color[HTML]{000000} }} \\ \cmidrule(r){1-5}
\multicolumn{1}{|l|}{{\color[HTML]{000000} SACrA}} & \multicolumn{1}{c|}{{\color[HTML]{000000} 8.64$^\uparrow$}} & \multicolumn{1}{c|}{8.81$^\uparrow$} & \multicolumn{1}{c|}{{\color[HTML]{000000} 7.96}} & \multicolumn{1}{c|}{{\color[HTML]{000000} 8.47$^\uparrow$}} & {\color[HTML]{000000} \textbf{}} & \multicolumn{1}{l}{{\color[HTML]{000000} }} & \multicolumn{1}{l}{{\color[HTML]{000000} }} & \multicolumn{1}{l}{{\color[HTML]{000000} }} & \multicolumn{1}{l}{{\color[HTML]{000000} }} & {\color[HTML]{000000} } & \multicolumn{1}{l}{{\color[HTML]{000000} }} \\ \cmidrule(r){1-5}
\multicolumn{1}{|l|}{{\color[HTML]{000000} SACrA + UDISCAL}} & \multicolumn{1}{c|}{{\color[HTML]{000000} 8.23}} & \multicolumn{1}{c|}{8.54} & \multicolumn{1}{c|}{{\color[HTML]{000000} 7.95}} & \multicolumn{1}{c|}{{\color[HTML]{000000} 8.24}} & {\color[HTML]{000000} } & \multicolumn{1}{l}{{\color[HTML]{000000} }} & \multicolumn{1}{l}{{\color[HTML]{000000} }} & \multicolumn{1}{l}{{\color[HTML]{000000} }} & \multicolumn{1}{l}{{\color[HTML]{000000} }} & {\color[HTML]{000000} } & \multicolumn{1}{l}{{\color[HTML]{000000} }} \\ \cmidrule(r){1-5}
\end{tabular}%
}
\caption{BLEU scores for the baseline Transformer model, previous work that used syntactically infused models -- PASCAL and UDISCAL, our SASA and SACrA models, and models incorporating UDISCAL with SASA or SACrA, across all WMT's newstests. For every language pair, each column contains the BLEU scores over the WMT newstest corresponding to the year the column is labeled with (e.g., for En-Ru, the scores under column \textit{2015} are for En-Ru newstest2015). For some newstests, there was more than one version on WMT, each translated by a different person. For those test sets, we included both versions, denoting the second one with a "B". In addition, for every language pair, the right-most column represents the average BLEU scores over all the pair's reported newstests. For every test set (and for the average score), the best score is boldfaced. For each of the semantic models (i.e., SASA and SACrA), improvements over all the baselines (syntactic and Transformer) are marked with an arrow facing upwards. For models with both syntactic and semantic masks, improvements over each mask individually are marked with an asterisk.}
\label{tab:full_testsets_Bleu}
\end{table*}

Next, we wish to compare our model to other baselines. Given that this is the first work to incorporate semantic information into the Transformer-based NMT model, we compare our work to syntactically-infused models (as described in \S\ref{sec:Baselines}): one is the PASCAL model \citep{bugliarello-okazaki-2020-enhancing}, and the other is our adaptation of PASCAL, the UD-Distance-Scaled (UDISCAL) model, which resembles better our SASA mask. We find (Table~\ref{tab:full_testsets_Bleu}) that on average, SASA outperforms both PASCAL and UDISCAL. We also compare SASA with each of the syntactic models, finding that it is significantly (sign-test $p<0.01$; see \S\ref{subsec:sign_test_full}) better. This suggests that semantics might be more beneficial for Transformers than syntax.

\subsection{Combining Syntax and Semantics}

Naturally, our next question is whether combining both semantic and syntactic heads will further improve the model's performance. Therefore, we test the combination of SASA with either PASCAL or UDISCAL, retaining the hyperparameters used for the separate models. We find that combining with UDISCAL outperforms the former, and so we continue with it. Interestingly, En-De and En-Ru hardly benefit from the combination compared just to the SASA model. We hypothesize that this might be due to the fact that the syntax of each language pair is already quite similar, and therefore the model mainly relies on it to separate the sentence that UCCA gives it as well. On the other hand, En-Fi and En-Tr do benefit from the combination, both on average and in most of the test sets. Evaluating the performance using the chrF metric (see \S\ref{subsec:ChrF_Results}) yields a similar behavior, which further confirms its validity. It leads us to hypothesize that language pairs that are more typologically distant from one another can benefit more from both semantics and syntax; we defer a more complete discussion of this point to future work. In order to confirm that the combined version persistently outperforms each of the separate versions for typologically distant languages, we compare each of the pairs using the sign-test (only on the test sets of En-Fi and En-Tr). We get a p-value of 0.02 for the comparison with SASA and 0.0008 for the comparison with UDISCAL. This suggests that for these language pairs, there is indeed a significant benefit, albeit small, from the infusion of both semantics and syntax.


\subsection{Scene-Aware Cross-Attention}\label{subsec:Scene-Aware Cross-Attention - Experiments}

Following the analysis on the scene-aware \textit{self}-attention, we wish to examine whether  Transformers could also benefit from injecting source-side semantics into the decoder. For that, we develop the Scene-Aware Cross-Attention (SACrA) model, as described in \S\ref{subsec:Same-Scene-Aware Cross-Attention}. Table \ref{tab:full_testsets_Bleu} presents the results of SACrA, compared to the Transformer baseline and SASA. We find that in general SASA outperforms SACrA, suggesting that semantics is more beneficial during encoding. With that said, for three out of the four language pairs,
SACrA does yield gains over the Transformer, albeit small, and for one language pair (En-Fi) it even outperforms SASA on average. Moreover, comparing SACrA to the Transformer using the sign-test (see \S\ref{subsec:sign_test_full}) shows significant improvement ($p=0.047$).

Surprisingly, unlike its self-attention counterpart, combining the SACrA model with UDISCAL does not seem to be beneficial at all, and in most cases is even outperformed by the baseline Transformer. We hypothesize that this occurs because appointing too many heads for our linguistic injection is inefficient when those heads cannot interact with each other directly, as the information from the UDISCAL head reaches the SACrA head only after the encoding is done. One possible direction for future work would be to find ways to syntactically enrich the decoder, and then to combine it with our SACrA model.  

\section{Conclusion}
 In this work, we suggest two novel methods for injecting semantic information into an NMT Transformer model  -- one through the encoder (i.e. SASA) and one through the decoder (i.e. SACrA). The strength of both methods is that they both do not introduce more parameters to the model, and only rely on UCCA-parses of the source sentences, which are generated in advance using an off-the-shelf parser, and thus do not increase the complexity of the model. 
 We compare our methods to previously developed methods of syntax injection, and to our adaptation to these methods, and find that semantic information tends to be significantly more beneficial than syntactic information, mostly when injected into the encoder (SASA), but at times also during decoding (SACrA). Moreover, we find that for sufficiently different languages, such as English and Finnish or English and Turkish, incorporating both syntactic and semantic structures further improves the performance of the translation models. Future work will further investigate the benefits of semantic structure in Transformers, alone and in unison with syntactic structure.
 
\section*{Acknowledgments}
This work was supported in part by the Israel Science Foundation (grant no. 2424/21), and by the Applied Research in Academia Program of the Israel Innovation Authority.


\bibliography{anthology, references}
\bibliographystyle{acl_natbib}

\clearpage
\pdfoutput=1

\appendix
\section{Appendix}\label{sec:appendix}
    \subsection{Layer Hyperparameter-tuning for SASA}\label{subsec:SASA_Layer_Hypertuning}
        In order to optimize the contribution of the SASA model, we tuned the hyperparameter of the best layers in the encoder to incorporate our model, using the En-De newstest2013 as our development set. Table \ref{tab:layer_hypertuning_SASA} presents the results.
    \subsection{ChrF Results}\label{subsec:ChrF_Results}
        In order to reaffirm our results, we also evaluate the performance of all the models using the chrF metric (see \ref{tab:chrf_results}). Indeed, all the different behaviors and trends we observed when evaluating using the Bleu metric (see \S\ref{sec:Experiments}) seem to be preserved when under the chrF metric. This further validates our results. 
    
    \subsection{Challenge Sets}\label{subsec:Challenge_Sets}
        In addition to testing on the full newstests sets, we also experiment with sentences characterized by long dependencies, which were shown to present a challenge for Transformers \citep{choshen-abend-2019-automatically}. In order to acquire those challenge sets, we use the methodology described by \citet{choshen-abend-2019-automatically}, which we apply on each of the newstest sets.
        In addition, for the En-Tr task, which has a limited number of newstests, we generate additional challenge sets, extracted from corpora downloaded from the Opus Corpus engine \citep{TIEDEMANN12.463}:  the Wikipedia parallel corpus \citep{WOLK2014126}, the Mozilla and EUbookshop parallel corpora \citep{TIEDEMANN12.463}, and the bible parallel corpus \citep{DBLP:journals/lre/Christodoulopoulos15}. 
        We observe (see \ref{tab:challenge_sets_Bleu}) a similar trend to the general case, which reaffirms our results. In fact, there seem to be bigger gains over the Transformer, albeit not drastically, compared to the general case.
    
    \subsection{Sign-Test}\label{subsec:sign_test_full}
        In order to assess the consistency of the improvements of our models, we perform the Sign-Test on every two models (see \ref{tab:sign_test_all}). Evidently, SASA persistently outperforms the Transformer baseline and the syntactic models, as does the combined model of SASA and UDISCAL.
        \begin{table}[t!]
\resizebox{0.3\columnwidth}{!}{%
\begin{tabular}{@{}|l|c|@{}}
\toprule
Layers & Bleu \\ \midrule
1 & 20.3 \\ \midrule
2 & 20.33 \\ \midrule
3 & 20.1 \\ \midrule
4 & 20.37 \\ \midrule
1,2 & 20.2 \\ \midrule
2,3 & 20.17 \\ \midrule
3,4 & 20.3 \\ \bottomrule
\end{tabular}%
}
\caption{Validation Bleu as a function of layers incorporating SASA (for En-De).}
\label{tab:layer_hypertuning_SASA}
\end{table}


\begin{table}[!]
\resizebox{\columnwidth}{!}{%
\begin{tabular}{@{}lcccccc@{}}
\backslashbox{\textbf{BASELINE}}{\textbf{BETTER}} & \multicolumn{1}{l}{\textbf{PASCAL}} & \multicolumn{1}{l}{\textbf{UDISCAL}} & \multicolumn{1}{l}{\textbf{SASA}} & \multicolumn{1}{l}{\textbf{\begin{tabular}[c]{@{}l@{}}SASA\\ + UDISCAL\end{tabular}}} & \multicolumn{1}{l}{\textbf{SACrA}} & \multicolumn{1}{l}{\textbf{\begin{tabular}[c]{@{}l@{}}SACrA\\ + UDISCAL\end{tabular}}} \\ [0.5cm]
\fontsize{13}{7.2}\selectfont \textbf{Transformer} & \fontsize{15}{7.2}\selectfont >0.5 & \fontsize{15}{7.2}\selectfont >0.5 & \fontsize{15}{7.2}\selectfont <0.01 & \fontsize{15}{7.2}\selectfont <0.01 & \fontsize{15}{7.2}\selectfont 0.047 & \fontsize{15}{7.2}\selectfont >0.5 \\
\textbf{PASCAL} &  & \fontsize{15}{7.2}\selectfont 0.17 & \fontsize{15}{7.2}\selectfont <0.01 & \fontsize{15}{7.2}\selectfont <0.01 & \fontsize{15}{7.2}\selectfont 0.06 & \fontsize{15}{7.2}\selectfont >0.5 \\
\textbf{UDISCAL} &  &  & \fontsize{15}{7.2}\selectfont <0.01 & \fontsize{15}{7.2}\selectfont <0.01 & \fontsize{15}{7.2}\selectfont 0.06 & \fontsize{15}{7.2}\selectfont >0.5 \\
\textbf{SASA} &  &  &  & \fontsize{15}{7.2}\selectfont 0.17 & \fontsize{15}{7.2}\selectfont >0.5 & \fontsize{15}{7.2}\selectfont >0.5 \\
\textbf{SASA + UDISCAL} &  &  &  &  & \fontsize{15}{7.2}\selectfont >0.5 & \fontsize{15}{7.2}\selectfont >0.5 \\
\textbf{SACrA} &  &  &  &  &  & \fontsize{15}{7.2}\selectfont >0.5
\end{tabular}%
}
\caption{We perform a significance test over all test sets across all languages for every cell, where the null hypothesis is $H_0: Bleu(model_{row}) \geq Bleu(model_{column})$}
\label{tab:sign_test_all}
\end{table}
        \begin{table*}[t!]
\resizebox{\textwidth}{!}{%
\begin{tabular}{@{}clcccccccclll@{}}
\multicolumn{12}{c}{{\ul \textbf{En-De}}} &  \\ \cmidrule(r){1-12}
\multicolumn{1}{|l|}{\textbf{Metric}} & \multicolumn{1}{l|}{\textbf{Models}} & \multicolumn{1}{c|}{\textbf{2012}} & \multicolumn{1}{c|}{\textbf{2014}} & \multicolumn{1}{c|}{\textbf{2015}} & \multicolumn{1}{c|}{\textbf{2016}} & \multicolumn{1}{c|}{\textbf{2017}} & \multicolumn{1}{c|}{\textbf{2018}} & \multicolumn{1}{c|}{\textbf{2019}} & \multicolumn{1}{c|}{\textbf{2020}} & \multicolumn{1}{c|}{\textbf{2020B}} & \multicolumn{1}{c|}{{\ul \textbf{average}}} &  \\ \cmidrule(r){1-12}
\multicolumn{1}{|c|}{} & \multicolumn{1}{l|}{Transformer} & \multicolumn{1}{c|}{17.6} & \multicolumn{1}{c|}{20.55} & \multicolumn{1}{c|}{22.17} & \multicolumn{1}{c|}{25.46} & \multicolumn{1}{c|}{19.7} & \multicolumn{1}{c|}{28.01} & \multicolumn{1}{c|}{26.84} & \multicolumn{1}{c|}{17.71} & \multicolumn{1}{c|}{16.94} & \multicolumn{1}{c|}{21.66} &  \\ \cmidrule(lr){2-12}
\multicolumn{1}{|c|}{\multirow{-2}{*}{\textbf{Bleu}}} & \multicolumn{1}{l|}{SemSplit} & \multicolumn{1}{c|}{12.16} & \multicolumn{1}{c|}{14.25} & \multicolumn{1}{c|}{14.46} & \multicolumn{1}{c|}{17.53} & \multicolumn{1}{c|}{13.18} & \multicolumn{1}{c|}{19.39} & \multicolumn{1}{c|}{18.46} & \multicolumn{1}{c|}{15.12} & \multicolumn{1}{c|}{14.93} & \multicolumn{1}{c|}{15.50} &  \\ \cmidrule(r){1-12}
\multicolumn{1}{|c|}{} & \multicolumn{1}{l|}{Transformer} & \multicolumn{1}{c|}{47.37} & \multicolumn{1}{c|}{51.85} & \multicolumn{1}{c|}{52.52} & \multicolumn{1}{c|}{55.06} & \multicolumn{1}{c|}{50.87} & \multicolumn{1}{c|}{57.81} & \multicolumn{1}{c|}{55.48} & \multicolumn{1}{c|}{45.19} & \multicolumn{1}{c|}{44.18} & \multicolumn{1}{c|}{51.15} &  \\ \cmidrule(lr){2-12}
\multicolumn{1}{|c|}{\multirow{-2}{*}{\textbf{chrF}}} & \multicolumn{1}{l|}{SemSplit} & \multicolumn{1}{c|}{43.42} & \multicolumn{1}{c|}{47.19} & \multicolumn{1}{c|}{47.05} & \multicolumn{1}{c|}{49.86} & \multicolumn{1}{c|}{45.87} & \multicolumn{1}{c|}{51.50} & \multicolumn{1}{c|}{50.24} & \multicolumn{1}{c|}{47.71} & \multicolumn{1}{c|}{46.93} & \multicolumn{1}{c|}{47.75} &  \\ \cmidrule(r){1-12}
\multicolumn{13}{c}{{\ul \textbf{En-Ru}}} \\ \midrule
\multicolumn{1}{|l|}{\textbf{Metric}} & \multicolumn{1}{l|}{\textbf{Models}} & \multicolumn{1}{c|}{\textbf{2012}} & \multicolumn{1}{c|}{\textbf{2013}} & \multicolumn{1}{c|}{\textbf{2014}} & \multicolumn{1}{c|}{\textbf{2015}} & \multicolumn{1}{c|}{\textbf{2016}} & \multicolumn{1}{c|}{\textbf{2017}} & \multicolumn{1}{c|}{\textbf{2018}} & \multicolumn{1}{c|}{\textbf{2019}} & \multicolumn{1}{c|}{\textbf{2020}} & \multicolumn{1}{c|}{\textbf{2020B}} & \multicolumn{1}{c|}{{\ul \textbf{average}}} \\ \midrule
\multicolumn{1}{|c|}{} & \multicolumn{1}{l|}{Transformer} & \multicolumn{1}{c|}{24.32} & \multicolumn{1}{c|}{18.11} & \multicolumn{1}{c|}{25.35} & \multicolumn{1}{c|}{21.1} & \multicolumn{1}{c|}{19.77} & \multicolumn{1}{c|}{22.34} & \multicolumn{1}{c|}{19} & \multicolumn{1}{c|}{20.14} & \multicolumn{1}{c|}{15.64} & \multicolumn{1}{c|}{22.33} & \multicolumn{1}{c|}{20.81} \\ \cmidrule(l){2-13} 
\multicolumn{1}{|c|}{\multirow{-2}{*}{\textbf{Bleu}}} & \multicolumn{1}{l|}{SemSplit} & \multicolumn{1}{c|}{15.29} & \multicolumn{1}{c|}{10.9} & \multicolumn{1}{c|}{16.43} & \multicolumn{1}{c|}{13.28} & \multicolumn{1}{c|}{12.79} & \multicolumn{1}{c|}{14.61} & \multicolumn{1}{c|}{11.95} & \multicolumn{1}{c|}{12.56} & \multicolumn{1}{c|}{9.92} & \multicolumn{1}{c|}{15.25} & \multicolumn{1}{c|}{13.30} \\ \midrule
\multicolumn{1}{|c|}{} & \multicolumn{1}{l|}{Transformer} & \multicolumn{1}{c|}{51.39} & \multicolumn{1}{c|}{45.69} & \multicolumn{1}{c|}{53.31} & \multicolumn{1}{c|}{50.16} & \multicolumn{1}{c|}{48.10} & \multicolumn{1}{c|}{50.54} & \multicolumn{1}{c|}{48.01} & \multicolumn{1}{c|}{45.78} & \multicolumn{1}{c|}{42.51} & \multicolumn{1}{c|}{53.07} & \multicolumn{1}{c|}{48.86} \\ \cmidrule(l){2-13} 
\multicolumn{1}{|c|}{\multirow{-2}{*}{\textbf{chrF}}} & \multicolumn{1}{l|}{SemSplit} & \multicolumn{1}{c|}{46.10} & \multicolumn{1}{c|}{40.50} & \multicolumn{1}{c|}{47.66} & \multicolumn{1}{c|}{44.58} & \multicolumn{1}{c|}{43.16} & \multicolumn{1}{c|}{45.34} & \multicolumn{1}{c|}{43.38} & \multicolumn{1}{c|}{40.97} & \multicolumn{1}{c|}{38.93} & \multicolumn{1}{c|}{47.84} & \multicolumn{1}{c|}{43.85} \\ \midrule
\multicolumn{1}{l}{} & \multicolumn{9}{c}{\cellcolor[HTML]{FFFFFF}{\ul \textbf{En-Fi}}} &  &  &  \\ \cmidrule(r){1-10}
\multicolumn{1}{|l|}{\textbf{Metric}} & \multicolumn{1}{l|}{\textbf{Models}} & \multicolumn{1}{c|}{\textbf{2015}} & \multicolumn{1}{c|}{\textbf{2016}} & \multicolumn{1}{c|}{\textbf{2016B}} & \multicolumn{1}{c|}{\textbf{2017}} & \multicolumn{1}{c|}{\textbf{2017B}} & \multicolumn{1}{c|}{\textbf{2018}} & \multicolumn{1}{c|}{\textbf{2019}} & \multicolumn{1}{c|}{{\ul \textbf{average}}} &  &  &  \\ \cmidrule(r){1-10}
\multicolumn{1}{|c|}{} & \multicolumn{1}{l|}{Transformer} & \multicolumn{1}{c|}{11.22} & \multicolumn{1}{c|}{12.76} & \multicolumn{1}{c|}{10.2} & \multicolumn{1}{c|}{13.35} & \multicolumn{1}{c|}{11.37} & \multicolumn{1}{c|}{9.32} & \multicolumn{1}{c|}{12.21} & \multicolumn{1}{c|}{11.49} &  &  &  \\ \cmidrule(lr){2-10}
\multicolumn{1}{|c|}{\multirow{-2}{*}{\textbf{Bleu}}} & \multicolumn{1}{l|}{SemSplit} & \multicolumn{1}{c|}{6.97} & \multicolumn{1}{c|}{7.72} & \multicolumn{1}{c|}{6.55} & \multicolumn{1}{c|}{8.75} & \multicolumn{1}{c|}{7.54} & \multicolumn{1}{c|}{6.18} & \multicolumn{1}{c|}{7.73} & \multicolumn{1}{c|}{7.35} &  &  &  \\ \cmidrule(r){1-10}
\multicolumn{1}{|c|}{} & \multicolumn{1}{l|}{Transformer} & \multicolumn{1}{c|}{43.79} & \multicolumn{1}{c|}{45.48} & \multicolumn{1}{c|}{43.43} & \multicolumn{1}{c|}{46.39} & \multicolumn{1}{c|}{43.96} & \multicolumn{1}{c|}{42.06} & \multicolumn{1}{c|}{43.10} & \multicolumn{1}{c|}{44.03} &  &  &  \\ \cmidrule(lr){2-10}
\multicolumn{1}{|c|}{\multirow{-2}{*}{\textbf{chrF}}} & \multicolumn{1}{l|}{SemSplit} & \multicolumn{1}{c|}{40.18} & \multicolumn{1}{c|}{41.42} & \multicolumn{1}{c|}{39.94} & \multicolumn{1}{c|}{42.18} & \multicolumn{1}{c|}{40.20} & \multicolumn{1}{c|}{38.76} & \multicolumn{1}{c|}{40.12} & \multicolumn{1}{c|}{40.40} &  &  &  \\ \cmidrule(r){1-10}
\multicolumn{1}{l}{} & \multicolumn{5}{c}{\cellcolor[HTML]{FFFFFF}{\ul \textbf{En-Tr}}} &  &  &  &  & \multicolumn{1}{c}{} & \multicolumn{1}{c}{} &  \\ \cmidrule(r){1-6}
\multicolumn{1}{|l|}{\textbf{Metric}} & \multicolumn{1}{l|}{\textbf{Models}} & \multicolumn{1}{c|}{\textbf{2016}} & \multicolumn{1}{c|}{\textbf{2017}} & \multicolumn{1}{c|}{\textbf{2018}} & \multicolumn{1}{c|}{{\ul \textbf{average}}} & \multicolumn{1}{l}{} & \multicolumn{1}{l}{} & \multicolumn{1}{l}{} & \multicolumn{1}{l}{} &  &  &  \\ \cmidrule(r){1-6}
\multicolumn{1}{|c|}{} & \multicolumn{1}{l|}{Transformer} & \multicolumn{1}{c|}{8.43} & \multicolumn{1}{c|}{8.55} & \multicolumn{1}{c|}{8.1} & \multicolumn{1}{c|}{8.36} & \multicolumn{1}{l}{} & \multicolumn{1}{l}{} & \multicolumn{1}{l}{} & \multicolumn{1}{l}{} &  &  &  \\ \cmidrule(lr){2-6}
\multicolumn{1}{|c|}{\multirow{-2}{*}{\textbf{Bleu}}} & \multicolumn{1}{l|}{SemSplit} & \multicolumn{1}{c|}{6.15} & \multicolumn{1}{c|}{6.07} & \multicolumn{1}{c|}{5.37} & \multicolumn{1}{c|}{5.86} & \multicolumn{1}{l}{} & \multicolumn{1}{l}{} & \multicolumn{1}{l}{} & \multicolumn{1}{l}{} &  &  &  \\ \cmidrule(r){1-6}
\multicolumn{1}{|c|}{} & \multicolumn{1}{l|}{Transformer} & \multicolumn{1}{c|}{40.24} & \multicolumn{1}{c|}{40.37} & \multicolumn{1}{c|}{39.75} & \multicolumn{1}{c|}{40.12} & \multicolumn{1}{l}{} & \multicolumn{1}{l}{} & \multicolumn{1}{l}{} & \multicolumn{1}{l}{} &  &  &  \\ \cmidrule(lr){2-6}
\multicolumn{1}{|c|}{\multirow{-2}{*}{\textbf{chrF}}} & \multicolumn{1}{l|}{SemSplit} & \multicolumn{1}{c|}{39.04} & \multicolumn{1}{c|}{39.00} & \multicolumn{1}{c|}{38.85} & \multicolumn{1}{c|}{38.97} & \multicolumn{1}{l}{} & \multicolumn{1}{l}{} & \multicolumn{1}{l}{} & \multicolumn{1}{l}{} &  &  &  \\ \cmidrule(r){1-6}
\end{tabular}%
}
\caption{Bleu and ChrF scores of the baseline Transformer and the SemSplit model.}
\label{tab:semSplit_results}
\end{table*}
\begin{table*}[]
\resizebox{\textwidth}{!}{%
\begin{tabular}{@{}lccccccccccc@{}}
\rowcolor[HTML]{FFFFFF} 
\multicolumn{11}{c}{\cellcolor[HTML]{FFFFFF}{\ul \textbf{En-De}}} &  \\ \cmidrule(r){1-11}
\multicolumn{1}{|l|}{\textbf{models}} & \multicolumn{1}{c|}{\textbf{2012}} & \multicolumn{1}{c|}{\textbf{2014}} & \multicolumn{1}{c|}{\textbf{2015}} & \multicolumn{1}{c|}{\textbf{2016}} & \multicolumn{1}{c|}{\textbf{2017}} & \multicolumn{1}{c|}{\textbf{2018}} & \multicolumn{1}{c|}{\textbf{2019}} & \multicolumn{1}{c|}{\textbf{2020}} & \multicolumn{1}{c|}{\textbf{2020B}} & \multicolumn{1}{c|}{{\ul \textbf{average}}} & {\ul \textbf{}} \\ \cmidrule(r){1-11}
\multicolumn{1}{|l|}{Transformer} & \multicolumn{1}{c|}{{\color[HTML]{000000} 47.37}} & \multicolumn{1}{c|}{51.85} & \multicolumn{1}{c|}{{\color[HTML]{000000} 52.52}} & \multicolumn{1}{c|}{{\color[HTML]{000000} \textbf{55.06}}} & \multicolumn{1}{c|}{{\color[HTML]{000000} 50.87}} & \multicolumn{1}{c|}{{\color[HTML]{000000} 57.81}} & \multicolumn{1}{c|}{{\color[HTML]{000000} 55.48}} & \multicolumn{1}{c|}{{\color[HTML]{000000} \textbf{45.19}}} & \multicolumn{1}{c|}{{\color[HTML]{000000} \textbf{44.18}}} & \multicolumn{1}{c|}{{\color[HTML]{000000} 51.15}} & {\color[HTML]{000000} } \\ \cmidrule(r){1-11}
\multicolumn{1}{|l|}{PASCAL} & \multicolumn{1}{c|}{{\color[HTML]{000000} 47.27}} & \multicolumn{1}{c|}{51.87} & \multicolumn{1}{c|}{{\color[HTML]{000000} \textbf{52.82}}} & \multicolumn{1}{c|}{{\color[HTML]{000000} 54.73}} & \multicolumn{1}{c|}{{\color[HTML]{000000} 50.83}} & \multicolumn{1}{c|}{{\color[HTML]{000000} 57.65}} & \multicolumn{1}{c|}{{\color[HTML]{000000} 55.28}} & \multicolumn{1}{c|}{{\color[HTML]{000000} 44.80}} & \multicolumn{1}{c|}{{\color[HTML]{000000} 43.78}} & \multicolumn{1}{c|}{{\color[HTML]{000000} 51.00}} & {\color[HTML]{000000} } \\ \cmidrule(r){1-11}
\multicolumn{1}{|l|}{UDISCAL} & \multicolumn{1}{c|}{{\color[HTML]{000000} 47.26}} & \multicolumn{1}{c|}{51.95} & \multicolumn{1}{c|}{{\color[HTML]{000000} 52.45}} & \multicolumn{1}{c|}{{\color[HTML]{000000} 54.99}} & \multicolumn{1}{c|}{{\color[HTML]{000000} 50.78}} & \multicolumn{1}{c|}{{\color[HTML]{000000} 57.40}} & \multicolumn{1}{c|}{{\color[HTML]{000000} 55.30}} & \multicolumn{1}{c|}{{\color[HTML]{000000} 44.48}} & \multicolumn{1}{c|}{{\color[HTML]{000000} 43.43}} & \multicolumn{1}{c|}{{\color[HTML]{000000} 50.89}} & {\color[HTML]{000000} } \\ \cmidrule(r){1-11}
\multicolumn{1}{|l|}{SASA} & \multicolumn{1}{c|}{{\color[HTML]{000000} \textbf{47.48$^\uparrow$}}} & \multicolumn{1}{c|}{\textbf{52.03$^\uparrow$}} & \multicolumn{1}{c|}{{\color[HTML]{000000} 52.74}} & \multicolumn{1}{c|}{{\color[HTML]{000000} 54.99}} & \multicolumn{1}{c|}{{\color[HTML]{000000} \textbf{51.23$^\uparrow$}}} & \multicolumn{1}{c|}{{\color[HTML]{000000} \textbf{57.88$^\uparrow$}}} & \multicolumn{1}{c|}{{\color[HTML]{000000} \textbf{55.69$^\uparrow$}}} & \multicolumn{1}{c|}{{\color[HTML]{000000} 45.03}} & \multicolumn{1}{c|}{{\color[HTML]{000000} 43.99}} & \multicolumn{1}{c|}{{\color[HTML]{000000} \textbf{51.23$^\uparrow$}}} & {\color[HTML]{000000} } \\ \cmidrule(r){1-11}
\multicolumn{1}{|l|}{SASA + UDISCAL} & \multicolumn{1}{c|}{{\color[HTML]{000000} 47.42}} & \multicolumn{1}{c|}{51.94} & \multicolumn{1}{c|}{{\color[HTML]{000000} 52.50}} & \multicolumn{1}{c|}{\cellcolor[HTML]{FFFFFF}{\color[HTML]{000000} 55.00$^{*}$}} & \multicolumn{1}{c|}{\cellcolor[HTML]{FFFFFF}{\color[HTML]{000000} 50.86}} & \multicolumn{1}{c|}{{\color[HTML]{000000} 57.74}} & \multicolumn{1}{c|}{{\color[HTML]{000000} 55.62}} & \multicolumn{1}{c|}{{\color[HTML]{000000} 44.72}} & \multicolumn{1}{c|}{{\color[HTML]{000000} 43.62}} & \multicolumn{1}{c|}{{\color[HTML]{000000} 51.05}} & {\color[HTML]{000000} } \\ \cmidrule(r){1-11}
\multicolumn{1}{|l|}{SACrA} & \multicolumn{1}{c|}{{\color[HTML]{000000} 47.02}} & \multicolumn{1}{c|}{51.66} & \multicolumn{1}{c|}{{\color[HTML]{000000} 52.48}} & \multicolumn{1}{c|}{{\color[HTML]{000000} 54.49}} & \multicolumn{1}{c|}{{\color[HTML]{000000} 50.55}} & \multicolumn{1}{c|}{{\color[HTML]{000000} 57.16}} & \multicolumn{1}{c|}{{\color[HTML]{000000} 55.05}} & \multicolumn{1}{c|}{{\color[HTML]{000000} 44.08}} & \multicolumn{1}{c|}{{\color[HTML]{000000} 43.15}} & \multicolumn{1}{c|}{{\color[HTML]{000000} 50.63}} & {\color[HTML]{000000} } \\ \cmidrule(r){1-11}
\multicolumn{1}{|l|}{SACrA + UDISCAL} & \multicolumn{1}{c|}{{\color[HTML]{000000} 46.71}} & \multicolumn{1}{c|}{51.63} & \multicolumn{1}{c|}{{\color[HTML]{000000} 52.18}} & \multicolumn{1}{c|}{{\color[HTML]{000000} 54.37}} & \multicolumn{1}{c|}{{\color[HTML]{000000} 50.22}} & \multicolumn{1}{c|}{{\color[HTML]{000000} 57.20}} & \multicolumn{1}{c|}{{\color[HTML]{000000} 54.96}} & \multicolumn{1}{c|}{{\color[HTML]{000000} 43.42}} & \multicolumn{1}{c|}{{\color[HTML]{000000} 42.40}} & \multicolumn{1}{c|}{{\color[HTML]{000000} 50.34}} & {\color[HTML]{000000} } \\ \cmidrule(r){1-11}
\rowcolor[HTML]{FFFFFF} 
\multicolumn{12}{c}{\cellcolor[HTML]{FFFFFF}{\ul \textbf{En-Ru}}} \\ \midrule
\multicolumn{1}{|l|}{\textbf{models}} & \multicolumn{1}{c|}{\textbf{2012}} & \multicolumn{1}{c|}{\textbf{2013}} & \multicolumn{1}{c|}{\textbf{2014}} & \multicolumn{1}{c|}{\textbf{2015}} & \multicolumn{1}{c|}{\textbf{2016}} & \multicolumn{1}{c|}{\textbf{2017}} & \multicolumn{1}{c|}{\textbf{2018}} & \multicolumn{1}{c|}{\textbf{2019}} & \multicolumn{1}{c|}{\textbf{2020}} & \multicolumn{1}{c|}{\textbf{2020B}} & \multicolumn{1}{c|}{{\ul \textbf{average}}} \\ \midrule
\multicolumn{1}{|l|}{Transformer} & \multicolumn{1}{c|}{{\color[HTML]{000000} 51.39}} & \multicolumn{1}{c|}{45.69} & \multicolumn{1}{c|}{{\color[HTML]{000000} 53.31}} & \multicolumn{1}{c|}{{\color[HTML]{000000} 50.16}} & \multicolumn{1}{c|}{{\color[HTML]{000000} 48.10}} & \multicolumn{1}{c|}{{\color[HTML]{000000} 50.54}} & \multicolumn{1}{c|}{{\color[HTML]{000000} 48.01}} & \multicolumn{1}{c|}{{\color[HTML]{000000} 45.78}} & \multicolumn{1}{c|}{{\color[HTML]{000000} 42.51}} & \multicolumn{1}{c|}{{\color[HTML]{000000} 53.07}} & \multicolumn{1}{c|}{{\color[HTML]{000000} 48.86}} \\ \midrule
\multicolumn{1}{|l|}{PASCAL} & \multicolumn{1}{c|}{{\color[HTML]{000000} 51.03}} & \multicolumn{1}{c|}{45.66} & \multicolumn{1}{c|}{{\color[HTML]{000000} 53.04}} & \multicolumn{1}{c|}{{\color[HTML]{000000} 49.87}} & \multicolumn{1}{c|}{{\color[HTML]{000000} 48.05}} & \multicolumn{1}{c|}{{\color[HTML]{000000} 50.32}} & \multicolumn{1}{c|}{{\color[HTML]{000000} 47.98}} & \multicolumn{1}{c|}{{\color[HTML]{000000} 45.86}} & \multicolumn{1}{c|}{{\color[HTML]{000000} 42.35}} & \multicolumn{1}{c|}{{\color[HTML]{000000} 52.42}} & \multicolumn{1}{c|}{{\color[HTML]{000000} 48.66}} \\ \midrule
\multicolumn{1}{|l|}{UDISCAL} & \multicolumn{1}{c|}{{\color[HTML]{000000} 51.26}} & \multicolumn{1}{c|}{{\color[HTML]{000000} 45.73}} & \multicolumn{1}{c|}{{\color[HTML]{000000} 53.45}} & \multicolumn{1}{c|}{{\color[HTML]{000000} 50.01}} & \multicolumn{1}{c|}{{\color[HTML]{000000} 48.57}} & \multicolumn{1}{c|}{{\color[HTML]{000000} 50.50}} & \multicolumn{1}{c|}{{\color[HTML]{000000} 48.27}} & \multicolumn{1}{c|}{{\color[HTML]{000000} 46.03}} & \multicolumn{1}{c|}{{\color[HTML]{000000} 42.60}} & \multicolumn{1}{c|}{{\color[HTML]{000000} 52.89}} & \multicolumn{1}{c|}{{\color[HTML]{000000} 48.93}} \\ \midrule
\multicolumn{1}{|l|}{SASA} & \multicolumn{1}{c|}{{\color[HTML]{000000} 51.34}} & \multicolumn{1}{c|}{{\color[HTML]{000000} \textbf{45.81$^\uparrow$}}} & \multicolumn{1}{c|}{{\color[HTML]{000000} 53.49$^\uparrow$}} & \multicolumn{1}{c|}{{\color[HTML]{000000} \textbf{50.32$^\uparrow$}}} & \multicolumn{1}{c|}{{\color[HTML]{000000} \textbf{48.60$^\uparrow$}}} & \multicolumn{1}{c|}{{\color[HTML]{000000} 50.67$^\uparrow$}} & \multicolumn{1}{c|}{{\color[HTML]{000000} \textbf{48.45$^\uparrow$}}} & \multicolumn{1}{c|}{{\color[HTML]{000000} 45.81}} & \multicolumn{1}{c|}{{\color[HTML]{000000} 42.76$^\uparrow$}} & \multicolumn{1}{c|}{{\color[HTML]{000000} \textbf{53.62$^\uparrow$}}} & \multicolumn{1}{c|}{{\color[HTML]{000000} \textbf{49.09$^\uparrow$}}} \\ \midrule
\multicolumn{1}{|l|}{SASA + UDISCAL} & \multicolumn{1}{c|}{\cellcolor[HTML]{FFFFFF}{\color[HTML]{000000} \textbf{51.43$^{*}$}}} & \multicolumn{1}{c|}{45.67} & \multicolumn{1}{c|}{\cellcolor[HTML]{FFFFFF}{\color[HTML]{000000} \textbf{53.56$^{*}$}}} & \multicolumn{1}{c|}{\cellcolor[HTML]{FFFFFF}{\color[HTML]{000000} 50.03}} & \multicolumn{1}{c|}{\cellcolor[HTML]{FFFFFF}{\color[HTML]{000000} 48.29}} & \multicolumn{1}{c|}{\cellcolor[HTML]{FFFFFF}{\color[HTML]{000000} 50.67}} & \multicolumn{1}{c|}{\cellcolor[HTML]{FFFFFF}{\color[HTML]{000000} 48.25}} & \multicolumn{1}{c|}{\cellcolor[HTML]{FFFFFF}{\color[HTML]{000000} \textbf{46.08$^{*}$}}} & \multicolumn{1}{c|}{\cellcolor[HTML]{FFFFFF}{\color[HTML]{000000} \textbf{42.81$^{*}$}}} & \multicolumn{1}{c|}{{\color[HTML]{000000} 53.14}} & \multicolumn{1}{c|}{{\color[HTML]{000000} 48.99}} \\ \midrule
\multicolumn{1}{|l|}{SACrA} & \multicolumn{1}{c|}{{\color[HTML]{000000} 51.28}} & \multicolumn{1}{c|}{45.57} & \multicolumn{1}{c|}{{\color[HTML]{000000} 53.50$^\uparrow$}} & \multicolumn{1}{c|}{{\color[HTML]{000000} 49.81}} & \multicolumn{1}{c|}{{\color[HTML]{000000} 48.42}} & \multicolumn{1}{c|}{{\color[HTML]{000000} \textbf{50.82$^\uparrow$}}} & \multicolumn{1}{c|}{{\color[HTML]{000000} 48.28$^\uparrow$}} & \multicolumn{1}{c|}{{\color[HTML]{000000} 45.92}} & \multicolumn{1}{c|}{{\color[HTML]{000000} 42.68$^\uparrow$}} & \multicolumn{1}{c|}{{\color[HTML]{000000} 52.76}} & \multicolumn{1}{c|}{{\color[HTML]{000000} 48.90}} \\ \midrule
\multicolumn{1}{|l|}{SACrA + UDISCAL} & \multicolumn{1}{c|}{{\color[HTML]{000000} 50.58}} & \multicolumn{1}{c|}{45.31} & \multicolumn{1}{c|}{{\color[HTML]{000000} 52.90}} & \multicolumn{1}{c|}{{\color[HTML]{000000} 49.40}} & \multicolumn{1}{c|}{{\color[HTML]{000000} 47.77}} & \multicolumn{1}{c|}{{\color[HTML]{000000} 50.03}} & \multicolumn{1}{c|}{{\color[HTML]{000000} 47.49}} & \multicolumn{1}{c|}{{\color[HTML]{000000} 45.26}} & \multicolumn{1}{c|}{{\color[HTML]{000000} 42.33}} & \multicolumn{1}{c|}{{\color[HTML]{000000} 51.93}} & \multicolumn{1}{c|}{{\color[HTML]{000000} 48.30}} \\ \midrule
\multicolumn{9}{c}{\cellcolor[HTML]{FFFFFF}{\ul \textbf{En-Fi}}} & \cellcolor[HTML]{FFFFFF} &  &  \\ \cmidrule(r){1-9}
\multicolumn{1}{|l|}{\textbf{models}} & \multicolumn{1}{c|}{\textbf{2015}} & \multicolumn{1}{c|}{\textbf{2016}} & \multicolumn{1}{c|}{\textbf{2016B}} & \multicolumn{1}{c|}{\textbf{2017}} & \multicolumn{1}{c|}{\textbf{2017B}} & \multicolumn{1}{c|}{\textbf{2018}} & \multicolumn{1}{c|}{\textbf{2019}} & \multicolumn{1}{c|}{{\ul \textbf{average}}} & {\ul \textbf{}} & \textbf{} &  \\ \cmidrule(r){1-9}
\multicolumn{1}{|l|}{Transformer} & \multicolumn{1}{c|}{43.79} & \multicolumn{1}{c|}{45.48} & \multicolumn{1}{c|}{43.43} & \multicolumn{1}{c|}{46.39} & \multicolumn{1}{c|}{43.96} & \multicolumn{1}{c|}{42.06} & \multicolumn{1}{c|}{43.10} & \multicolumn{1}{c|}{44.03} &  &  &  \\ \cmidrule(r){1-9}
\multicolumn{1}{|l|}{PASCAL} & \multicolumn{1}{c|}{{\color[HTML]{000000} \textbf{43.91}}} & \multicolumn{1}{c|}{44.93} & \multicolumn{1}{c|}{{\color[HTML]{000000} 42.99}} & \multicolumn{1}{c|}{{\color[HTML]{000000} 46.02}} & \multicolumn{1}{c|}{{\color[HTML]{000000} 43.57}} & \multicolumn{1}{c|}{{\color[HTML]{000000} 41.88}} & \multicolumn{1}{c|}{{\color[HTML]{000000} 42.60}} & \multicolumn{1}{c|}{{\color[HTML]{000000} 43.70}} & {\color[HTML]{000000} } &  &  \\ \cmidrule(r){1-9}
\multicolumn{1}{|l|}{UDISCAL} & \multicolumn{1}{c|}{{\color[HTML]{000000} 43.42}} & \multicolumn{1}{c|}{45.37} & \multicolumn{1}{c|}{{\color[HTML]{000000} 43.42}} & \multicolumn{1}{c|}{{\color[HTML]{000000} 46.51}} & \multicolumn{1}{c|}{{\color[HTML]{000000} 44.07}} & \multicolumn{1}{c|}{{\color[HTML]{000000} 42.03}} & \multicolumn{1}{c|}{{\color[HTML]{000000} 43.03}} & \multicolumn{1}{c|}{{\color[HTML]{000000} 43.98}} & {\color[HTML]{000000} } &  &  \\ \cmidrule(r){1-9}
\multicolumn{1}{|l|}{SASA} & \multicolumn{1}{c|}{{\color[HTML]{000000} 43.76}} & \multicolumn{1}{c|}{45.33} & \multicolumn{1}{c|}{{\color[HTML]{000000} 43.38}} & \multicolumn{1}{c|}{{\color[HTML]{000000} 46.40}} & \multicolumn{1}{c|}{{\color[HTML]{000000} 43.89}} & \multicolumn{1}{c|}{{\color[HTML]{000000} 42.10$^\uparrow$}} & \multicolumn{1}{c|}{{\color[HTML]{000000} 43.02}} & \multicolumn{1}{c|}{{\color[HTML]{000000} 43.98}} & {\color[HTML]{000000} } &  &  \\ \cmidrule(r){1-9}
\multicolumn{1}{|l|}{SASA + UDISCAL} & \multicolumn{1}{c|}{\cellcolor[HTML]{FFFFFF}{\color[HTML]{000000} 43.77$^{*}$}} & \multicolumn{1}{c|}{\cellcolor[HTML]{FFFFFF}45.20} & \multicolumn{1}{c|}{\cellcolor[HTML]{FFFFFF}{\color[HTML]{000000} 43.17}} & \multicolumn{1}{c|}{\cellcolor[HTML]{FFFFFF}{\color[HTML]{000000} \textbf{46.74$^{*}$}}} & \multicolumn{1}{c|}{\cellcolor[HTML]{FFFFFF}{\color[HTML]{000000} 44.15$^{*}$}} & \multicolumn{1}{c|}{\cellcolor[HTML]{FFFFFF}{\color[HTML]{000000} \textbf{42.34$^{*}$}}} & \multicolumn{1}{c|}{\cellcolor[HTML]{FFFFFF}{\color[HTML]{000000} 43.08$^{*}$}} & \multicolumn{1}{c|}{\cellcolor[HTML]{FFFFFF}{\color[HTML]{000000} 44.07$^{*}$}} & \cellcolor[HTML]{FFFFFF}{\color[HTML]{000000} \textbf{}} &  &  \\ \cmidrule(r){1-9}
\multicolumn{1}{|l|}{SACrA} & \multicolumn{1}{c|}{{\color[HTML]{000000} 43.88}} & \multicolumn{1}{c|}{45.20} & \multicolumn{1}{c|}{{\color[HTML]{000000} 43.15}} & \multicolumn{1}{c|}{{\color[HTML]{000000} 46.62$^\uparrow$}} & \multicolumn{1}{c|}{{\color[HTML]{000000} 44.02$^\uparrow$}} & \multicolumn{1}{c|}{{\color[HTML]{000000} 42.25$^\uparrow$}} & \multicolumn{1}{c|}{{\color[HTML]{000000} 43.23$^\uparrow$}} & \multicolumn{1}{c|}{{\color[HTML]{000000} 44.05$^\uparrow$}} & {\color[HTML]{000000} \textbf{}} &  &  \\ \cmidrule(r){1-9}
\multicolumn{1}{|l|}{SACrA + UDISCAL} & \multicolumn{1}{c|}{\cellcolor[HTML]{FFFFFF}{\color[HTML]{000000} 43.80}} & \multicolumn{1}{c|}{\cellcolor[HTML]{FFFFFF}\textbf{45.53$^{*}$}} & \multicolumn{1}{c|}{\cellcolor[HTML]{FFFFFF}{\color[HTML]{000000} \textbf{43.52$^{*}$}}} & \multicolumn{1}{c|}{\cellcolor[HTML]{FFFFFF}{\color[HTML]{000000} 46.71$^{*}$}} & \multicolumn{1}{c|}{\cellcolor[HTML]{FFFFFF}{\color[HTML]{000000} \textbf{44.19$^{*}$}}} & \multicolumn{1}{c|}{\cellcolor[HTML]{FFFFFF}{\color[HTML]{000000} 42.16}} & \multicolumn{1}{c|}{\cellcolor[HTML]{FFFFFF}{\color[HTML]{000000} \textbf{43.28$^{*}$}}} & \multicolumn{1}{c|}{\cellcolor[HTML]{FFFFFF}{\color[HTML]{000000} \textbf{44.17$^{*}$}}} & \cellcolor[HTML]{FFFFFF}{\color[HTML]{000000} \textbf{}} &  &  \\ \cmidrule(r){1-9}
\multicolumn{5}{c}{\cellcolor[HTML]{FFFFFF}{\ul \textbf{En-Tr}}} & \cellcolor[HTML]{FFFFFF} &  &  &  &  &  &  \\ \cmidrule(r){1-5}
\multicolumn{1}{|l|}{{\color[HTML]{000000} \textbf{models}}} & \multicolumn{1}{c|}{{\color[HTML]{000000} \textbf{2016}}} & \multicolumn{1}{c|}{\textbf{2017}} & \multicolumn{1}{c|}{{\color[HTML]{000000} \textbf{2018}}} & \multicolumn{1}{c|}{{\color[HTML]{000000} {\ul \textbf{average}}}} & {\color[HTML]{000000} {\ul \textbf{}}} &  &  &  &  &  &  \\ \cmidrule(r){1-5}
\multicolumn{1}{|l|}{{\color[HTML]{000000} Transformer}} & \multicolumn{1}{c|}{{\color[HTML]{000000} 40.24}} & \multicolumn{1}{c|}{40.37} & \multicolumn{1}{c|}{{\color[HTML]{000000} 39.75}} & \multicolumn{1}{c|}{{\color[HTML]{000000} 40.12}} & {\color[HTML]{000000} } &  &  &  &  &  &  \\ \cmidrule(r){1-5}
\multicolumn{1}{|l|}{{\color[HTML]{000000} PASCAL}} & \multicolumn{1}{c|}{{\color[HTML]{000000} 40.59}} & \multicolumn{1}{c|}{40.64} & \multicolumn{1}{c|}{{\color[HTML]{000000} 39.89}} & \multicolumn{1}{c|}{{\color[HTML]{000000} 40.37}} & {\color[HTML]{000000} \textbf{}} &  &  &  &  &  &  \\ \cmidrule(r){1-5}
\multicolumn{1}{|l|}{{\color[HTML]{000000} UDISCAL}} & \multicolumn{1}{c|}{{\color[HTML]{000000} 40.27}} & \multicolumn{1}{c|}{40.49} & \multicolumn{1}{c|}{{\color[HTML]{000000} 40.01}} & \multicolumn{1}{c|}{{\color[HTML]{000000} 40.26}} & {\color[HTML]{000000} \textbf{}} &  &  &  &  &  &  \\ \cmidrule(r){1-5}
\multicolumn{1}{|l|}{{\color[HTML]{000000} SASA}} & \multicolumn{1}{c|}{{\color[HTML]{000000} 40.27}} & \multicolumn{1}{c|}{40.46} & \multicolumn{1}{c|}{{\color[HTML]{000000} 39.98}} & \multicolumn{1}{c|}{{\color[HTML]{000000} 40.24}} & {\color[HTML]{000000} \textbf{}} &  &  &  &  &  &  \\ \cmidrule(r){1-5}
\multicolumn{1}{|l|}{\cellcolor[HTML]{FFFFFF}{\color[HTML]{000000} SASA + UDISCAL}} & \multicolumn{1}{c|}{\cellcolor[HTML]{FFFFFF}{\color[HTML]{000000} \textbf{40.61$^{*}$}}} & \multicolumn{1}{c|}{\cellcolor[HTML]{FFFFFF}\textbf{40.92$^{*}$}} & \multicolumn{1}{c|}{\cellcolor[HTML]{FFFFFF}{\color[HTML]{000000} \textbf{40.12$^{*}$}}} & \multicolumn{1}{c|}{\cellcolor[HTML]{FFFFFF}{\color[HTML]{000000} \textbf{40.55$^{*}$}}} & {\color[HTML]{000000} \textbf{}} &  &  &  &  &  &  \\ \cmidrule(r){1-5}
\multicolumn{1}{|l|}{{\color[HTML]{000000} SACrA}} & \multicolumn{1}{c|}{{\color[HTML]{000000} 40.44}} & \multicolumn{1}{c|}{40.68$^\uparrow$} & \multicolumn{1}{c|}{{\color[HTML]{000000} 39.85}} & \multicolumn{1}{c|}{{\color[HTML]{000000} 40.33}} & {\color[HTML]{000000} \textbf{}} &  &  &  &  &  &  \\ \cmidrule(r){1-5}
\multicolumn{1}{|l|}{{\color[HTML]{000000} SACrA + UDISCAL}} & \multicolumn{1}{c|}{{\color[HTML]{000000} 40.23}} & \multicolumn{1}{c|}{40.48} & \multicolumn{1}{c|}{{\color[HTML]{000000} 39.96}} & \multicolumn{1}{c|}{{\color[HTML]{000000} 40.22}} & {\color[HTML]{000000} \textbf{}} &  &  &  &  &  &  \\ \cmidrule(r){1-5}
\end{tabular}%
}
\caption{ChrF scores for the baseline Transformer model, the baseline Syntactically infused models PASCAL and UDISCAL, our SASA and SACrA models, and models incorporating UDISCAL with each of SASA and SACrA, across all WMT's newstests. For every language pair, each column contains the Bleu scores over the WMT newstest equivalent to the column's year (e.g., for En-Ru, the scores under column \textit{2015} are for En-Ru newstest2015). For some newstests, there was more than one version on WMT, each translated by a different person. For those test sets, we included both versions, denoting the second one with a "B". In addition, for every language pair, the right-most column represents the average Bleu scores over all the pair's reported newstests. For every test set (and for the average score), the best score is boldfaced. For each of the semantic models (i.e., SASA and SACrA), improvements over all the baselines (syntactic and Transformer) are marked by an arrow facing upwards. For models with both syntactic and semantic masks, improvements over each mask individually are marked by an asterisk.}
\label{tab:chrf_results}
\end{table*}

\begin{table*}[]
\resizebox{\textwidth}{!}{%
\begin{tabular}{@{}lccccccccccc@{}}
\rowcolor[HTML]{FFFFFF} 
\multicolumn{11}{c}{\cellcolor[HTML]{FFFFFF}{\color[HTML]{000000} {\ul \textbf{En-De}}}} & {\color[HTML]{000000} {\ul \textbf{}}} \\ \cmidrule(r){1-11}
\multicolumn{1}{|l|}{{\color[HTML]{000000} \textbf{models}}} & \multicolumn{1}{c|}{{\color[HTML]{000000} \textbf{2012}}} & \multicolumn{1}{c|}{{\color[HTML]{000000} \textbf{2014}}} & \multicolumn{1}{c|}{{\color[HTML]{000000} \textbf{2015}}} & \multicolumn{1}{c|}{{\color[HTML]{000000} \textbf{2016}}} & \multicolumn{1}{c|}{{\color[HTML]{000000} \textbf{2017}}} & \multicolumn{1}{c|}{{\color[HTML]{000000} \textbf{2018}}} & \multicolumn{1}{c|}{{\color[HTML]{000000} \textbf{2019}}} & \multicolumn{1}{c|}{{\color[HTML]{000000} \textbf{2020}}} & \multicolumn{1}{c|}{{\color[HTML]{000000} \textbf{2020B}}} & \multicolumn{1}{c|}{{\color[HTML]{000000} {\ul \textbf{average}}}} & {\color[HTML]{000000} {\ul \textbf{}}} \\ \cmidrule(r){1-11}
\multicolumn{1}{|l|}{{\color[HTML]{000000} Transformer}} & \multicolumn{1}{c|}{{\color[HTML]{000000} 15.08}} & \multicolumn{1}{c|}{{\color[HTML]{000000} 16.94}} & \multicolumn{1}{c|}{{\color[HTML]{000000} 17.36}} & \multicolumn{1}{c|}{{\color[HTML]{000000} 21.11}} & \multicolumn{1}{c|}{{\color[HTML]{000000} 14.84}} & \multicolumn{1}{c|}{{\color[HTML]{000000} 23.43}} & \multicolumn{1}{c|}{{\color[HTML]{000000} 22.42}} & \multicolumn{1}{c|}{{\color[HTML]{000000} 16.79}} & \multicolumn{1}{c|}{{\color[HTML]{000000} 15.75}} & \multicolumn{1}{c|}{{\color[HTML]{000000} 18.19}} & {\color[HTML]{000000} } \\ \cmidrule(r){1-11}
\multicolumn{1}{|l|}{{\color[HTML]{000000} PASCAL}} & \multicolumn{1}{c|}{{\color[HTML]{000000} 14.96}} & \multicolumn{1}{c|}{{\color[HTML]{000000} 17.45}} & \multicolumn{1}{c|}{{\color[HTML]{000000} 17.85}} & \multicolumn{1}{c|}{{\color[HTML]{000000} 20.22}} & \multicolumn{1}{c|}{{\color[HTML]{000000} 14.66}} & \multicolumn{1}{c|}{{\color[HTML]{000000} 23.76}} & \multicolumn{1}{c|}{{\color[HTML]{000000} 21.28}} & \multicolumn{1}{c|}{{\color[HTML]{000000} \textbf{16.9}}} & \multicolumn{1}{c|}{{\color[HTML]{000000} \textbf{16.22}}} & \multicolumn{1}{c|}{{\color[HTML]{000000} 18.14}} & {\color[HTML]{000000} } \\ \cmidrule(r){1-11}
\multicolumn{1}{|l|}{{\color[HTML]{000000} UDISCAL}} & \multicolumn{1}{c|}{{\color[HTML]{000000} 14.46}} & \multicolumn{1}{c|}{{\color[HTML]{000000} \textbf{17.84}}} & \multicolumn{1}{c|}{{\color[HTML]{000000} 17.7}} & \multicolumn{1}{c|}{{\color[HTML]{000000} \textbf{21.26}}} & \multicolumn{1}{c|}{{\color[HTML]{000000} \textbf{15.48}}} & \multicolumn{1}{c|}{{\color[HTML]{000000} 23.75}} & \multicolumn{1}{c|}{{\color[HTML]{000000} 22.36}} & \multicolumn{1}{c|}{{\color[HTML]{000000} 16.37}} & \multicolumn{1}{c|}{{\color[HTML]{000000} 15.37}} & \multicolumn{1}{c|}{{\color[HTML]{000000} 18.29}} & {\color[HTML]{000000} \textbf{}} \\ \cmidrule(r){1-11}
\multicolumn{1}{|l|}{{\color[HTML]{000000} SASA}} & \multicolumn{1}{c|}{{\color[HTML]{000000} 14.67}} & \multicolumn{1}{c|}{{\color[HTML]{000000} 17.68}} & \multicolumn{1}{c|}{{\color[HTML]{000000} \textbf{18.04$^\uparrow$}}} & \multicolumn{1}{c|}{{\color[HTML]{000000} 20.89}} & \multicolumn{1}{c|}{{\color[HTML]{000000} 15.09}} & \multicolumn{1}{c|}{{\color[HTML]{000000} \textbf{24.8$^\uparrow$}}} & \multicolumn{1}{c|}{{\color[HTML]{000000} 22.86$^\uparrow$}} & \multicolumn{1}{c|}{{\color[HTML]{000000} 16.85}} & \multicolumn{1}{c|}{{\color[HTML]{000000} 15.76}} & \multicolumn{1}{c|}{{\color[HTML]{000000} \textbf{18.52$^\uparrow$}}} & {\color[HTML]{000000} \textbf{}} \\ \cmidrule(r){1-11}
\multicolumn{1}{|l|}{{\color[HTML]{000000} SASA + UDISCAL}} & \multicolumn{1}{c|}{\cellcolor[HTML]{FFFFFF}{\color[HTML]{000000} \textbf{15.39$^{*}$}}} & \multicolumn{1}{c|}{\cellcolor[HTML]{FFFFFF}{\color[HTML]{000000} 17.07}} & \multicolumn{1}{c|}{\cellcolor[HTML]{FFFFFF}{\color[HTML]{000000} 17.38}} & \multicolumn{1}{c|}{\cellcolor[HTML]{FFFFFF}{\color[HTML]{000000} 20.42}} & \multicolumn{1}{c|}{\cellcolor[HTML]{FFFFFF}{\color[HTML]{000000} 15.35}} & \multicolumn{1}{c|}{\cellcolor[HTML]{FFFFFF}{\color[HTML]{000000} 23.53}} & \multicolumn{1}{c|}{\cellcolor[HTML]{FFFFFF}{\color[HTML]{000000} \textbf{22.87$^{*}$}}} & \multicolumn{1}{c|}{\cellcolor[HTML]{FFFFFF}{\color[HTML]{000000} 16.79}} & \multicolumn{1}{c|}{\cellcolor[HTML]{FFFFFF}{\color[HTML]{000000} 15.98$^{*}$}} & \multicolumn{1}{c|}{\cellcolor[HTML]{FFFFFF}{\color[HTML]{000000} 18.31}} & {\color[HTML]{000000} \textbf{}} \\ \cmidrule(r){1-11}
\multicolumn{1}{|l|}{{\color[HTML]{000000} SACrA}} & \multicolumn{1}{c|}{{\color[HTML]{000000} 14.67}} & \multicolumn{1}{c|}{{\color[HTML]{000000} 17.03}} & \multicolumn{1}{c|}{{\color[HTML]{000000} 16.89}} & \multicolumn{1}{c|}{{\color[HTML]{000000} 19.69}} & \multicolumn{1}{c|}{{\color[HTML]{000000} 14.45}} & \multicolumn{1}{c|}{{\color[HTML]{000000} 22.21}} & \multicolumn{1}{c|}{{\color[HTML]{000000} 22.08}} & \multicolumn{1}{c|}{{\color[HTML]{000000} 16.64}} & \multicolumn{1}{c|}{{\color[HTML]{000000} 15.6}} & \multicolumn{1}{c|}{{\color[HTML]{000000} 17.70}} & {\color[HTML]{000000} } \\ \cmidrule(r){1-11}
\multicolumn{1}{|l|}{{\color[HTML]{000000} SACrA + UDISCAL}} & \multicolumn{1}{c|}{{\color[HTML]{000000} 15.07$^{*}$}} & \multicolumn{1}{c|}{{\color[HTML]{000000} 17.23}} & \multicolumn{1}{c|}{{\color[HTML]{000000} 16.52}} & \multicolumn{1}{c|}{{\color[HTML]{000000} 20.82}} & \multicolumn{1}{c|}{{\color[HTML]{000000} 14.6}} & \multicolumn{1}{c|}{{\color[HTML]{000000} 22.38}} & \multicolumn{1}{c|}{{\color[HTML]{000000} 22.61$^{*}$}} & \multicolumn{1}{c|}{{\color[HTML]{000000} 16.53}} & \multicolumn{1}{c|}{{\color[HTML]{000000} 15.81$^{*}$}} & \multicolumn{1}{c|}{{\color[HTML]{000000} 17.95}} & {\color[HTML]{000000} } \\ \cmidrule(r){1-11}
\rowcolor[HTML]{FFFFFF} 
\multicolumn{12}{c}{\cellcolor[HTML]{FFFFFF}{\color[HTML]{000000} {\ul \textbf{En-Ru}}}} \\ \midrule
\multicolumn{1}{|l|}{{\color[HTML]{000000} \textbf{models}}} & \multicolumn{1}{c|}{{\color[HTML]{000000} \textbf{2012}}} & \multicolumn{1}{c|}{{\color[HTML]{000000} \textbf{2013}}} & \multicolumn{1}{c|}{{\color[HTML]{000000} \textbf{2014}}} & \multicolumn{1}{c|}{{\color[HTML]{000000} \textbf{2015}}} & \multicolumn{1}{c|}{{\color[HTML]{000000} \textbf{2016}}} & \multicolumn{1}{c|}{{\color[HTML]{000000} \textbf{2017}}} & \multicolumn{1}{c|}{{\color[HTML]{000000} \textbf{2018}}} & \multicolumn{1}{c|}{{\color[HTML]{000000} \textbf{2019}}} & \multicolumn{1}{c|}{{\color[HTML]{000000} \textbf{2020}}} & \multicolumn{1}{c|}{{\color[HTML]{000000} \textbf{2020B}}} & \multicolumn{1}{c|}{{\color[HTML]{000000} {\ul \textbf{average}}}} \\ \midrule
\multicolumn{1}{|l|}{{\color[HTML]{000000} Transformer}} & \multicolumn{1}{c|}{\cellcolor[HTML]{FFFFFF}{\color[HTML]{000000} 23.4}} & \multicolumn{1}{c|}{\cellcolor[HTML]{FFFFFF}{\color[HTML]{000000} 14.67}} & \multicolumn{1}{c|}{\cellcolor[HTML]{FFFFFF}{\color[HTML]{000000} \textbf{24}}} & \multicolumn{1}{c|}{\cellcolor[HTML]{FFFFFF}{\color[HTML]{000000} 16.82}} & \multicolumn{1}{c|}{\cellcolor[HTML]{FFFFFF}{\color[HTML]{000000} 17.52}} & \multicolumn{1}{c|}{\cellcolor[HTML]{FFFFFF}{\color[HTML]{000000} 19.74}} & \multicolumn{1}{c|}{\cellcolor[HTML]{FFFFFF}{\color[HTML]{000000} 17.78}} & \multicolumn{1}{c|}{\cellcolor[HTML]{FFFFFF}{\color[HTML]{000000} 17.12}} & \multicolumn{1}{c|}{\cellcolor[HTML]{FFFFFF}{\color[HTML]{000000} 13.39}} & \multicolumn{1}{c|}{{\color[HTML]{000000} 19.47}} & \multicolumn{1}{c|}{{\color[HTML]{000000} 18.39}} \\ \midrule
\multicolumn{1}{|l|}{{\color[HTML]{000000} PASCAL}} & \multicolumn{1}{c|}{\cellcolor[HTML]{FFFFFF}{\color[HTML]{000000} 22.6}} & \multicolumn{1}{c|}{\cellcolor[HTML]{FFFFFF}{\color[HTML]{000000} \textbf{15.67}}} & \multicolumn{1}{c|}{\cellcolor[HTML]{FFFFFF}{\color[HTML]{000000} 23.56}} & \multicolumn{1}{c|}{\cellcolor[HTML]{FFFFFF}{\color[HTML]{000000} 17.08}} & \multicolumn{1}{c|}{\cellcolor[HTML]{FFFFFF}{\color[HTML]{000000} 17.79}} & \multicolumn{1}{c|}{\cellcolor[HTML]{FFFFFF}{\color[HTML]{000000} 19.46}} & \multicolumn{1}{c|}{\cellcolor[HTML]{FFFFFF}{\color[HTML]{000000} 17.9}} & \multicolumn{1}{c|}{\cellcolor[HTML]{FFFFFF}{\color[HTML]{000000} 16.13}} & \multicolumn{1}{c|}{\cellcolor[HTML]{FFFFFF}{\color[HTML]{000000} 13.7}} & \multicolumn{1}{c|}{{\color[HTML]{000000} 19.44}} & \multicolumn{1}{c|}{{\color[HTML]{000000} 18.33}} \\ \midrule
\multicolumn{1}{|l|}{{\color[HTML]{000000} UDISCAL}} & \multicolumn{1}{c|}{\cellcolor[HTML]{FFFFFF}{\color[HTML]{000000} 23.19}} & \multicolumn{1}{c|}{\cellcolor[HTML]{FFFFFF}{\color[HTML]{000000} 14.75}} & \multicolumn{1}{c|}{\cellcolor[HTML]{FFFFFF}{\color[HTML]{000000} 23.46}} & \multicolumn{1}{c|}{\cellcolor[HTML]{FFFFFF}{\color[HTML]{000000} 17.06}} & \multicolumn{1}{c|}{\cellcolor[HTML]{FFFFFF}{\color[HTML]{000000} 18.17}} & \multicolumn{1}{c|}{\cellcolor[HTML]{FFFFFF}{\color[HTML]{000000} 19.67}} & \multicolumn{1}{c|}{\cellcolor[HTML]{FFFFFF}{\color[HTML]{000000} 18.32}} & \multicolumn{1}{c|}{\cellcolor[HTML]{FFFFFF}{\color[HTML]{000000} 15.7}} & \multicolumn{1}{c|}{\cellcolor[HTML]{FFFFFF}{\color[HTML]{000000} 13.44}} & \multicolumn{1}{c|}{{\color[HTML]{000000} \textbf{21.14}}} & \multicolumn{1}{c|}{{\color[HTML]{000000} 18.49}} \\ \midrule
\multicolumn{1}{|l|}{{\color[HTML]{000000} SASA}} & \multicolumn{1}{c|}{\cellcolor[HTML]{FFFFFF}{\color[HTML]{000000} 23.53$^\uparrow$}} & \multicolumn{1}{c|}{\cellcolor[HTML]{FFFFFF}{\color[HTML]{000000} 15.38}} & \multicolumn{1}{c|}{\cellcolor[HTML]{FFFFFF}{\color[HTML]{000000} 23.9}} & \multicolumn{1}{c|}{\cellcolor[HTML]{FFFFFF}{\color[HTML]{000000} 17.77$^\uparrow$}} & \multicolumn{1}{c|}{\cellcolor[HTML]{FFFFFF}{\color[HTML]{000000} \textbf{18.37$^\uparrow$}}} & \multicolumn{1}{c|}{\cellcolor[HTML]{FFFFFF}{\color[HTML]{000000} \textbf{20.12$^\uparrow$}}} & \multicolumn{1}{c|}{\cellcolor[HTML]{FFFFFF}{\color[HTML]{000000} 18.33$^\uparrow$}} & \multicolumn{1}{c|}{\cellcolor[HTML]{FFFFFF}{\color[HTML]{000000} 16.55}} & \multicolumn{1}{c|}{\cellcolor[HTML]{FFFFFF}{\color[HTML]{000000} 13.37}} & \multicolumn{1}{c|}{{\color[HTML]{000000} 20.88}} & \multicolumn{1}{c|}{{\color[HTML]{000000} \textbf{18.82$^\uparrow$}}} \\ \midrule
\multicolumn{1}{|l|}{{\color[HTML]{000000} SASA + UDISCAL}} & \multicolumn{1}{c|}{\cellcolor[HTML]{FFFFFF}{\color[HTML]{000000} 23.77$^{*}$}} & \multicolumn{1}{c|}{\cellcolor[HTML]{FFFFFF}{\color[HTML]{000000} 14.67}} & \multicolumn{1}{c|}{\cellcolor[HTML]{FFFFFF}{\color[HTML]{000000} 23.65}} & \multicolumn{1}{c|}{\cellcolor[HTML]{FFFFFF}{\color[HTML]{000000} 16.96}} & \multicolumn{1}{c|}{\cellcolor[HTML]{FFFFFF}{\color[HTML]{000000} 18.21}} & \multicolumn{1}{c|}{\cellcolor[HTML]{FFFFFF}{\color[HTML]{000000} 19.8}} & \multicolumn{1}{c|}{\cellcolor[HTML]{FFFFFF}{\color[HTML]{000000} 18.06}} & \multicolumn{1}{c|}{\cellcolor[HTML]{FFFFFF}{\color[HTML]{000000} \textbf{17.15$^{*}$}}} & \multicolumn{1}{c|}{\cellcolor[HTML]{FFFFFF}{\color[HTML]{000000} 13.57$^{*}$}} & \multicolumn{1}{c|}{\cellcolor[HTML]{FFFFFF}{\color[HTML]{000000} 20.02}} & \multicolumn{1}{c|}{\cellcolor[HTML]{FFFFFF}{\color[HTML]{000000} 18.59}} \\ \midrule
\rowcolor[HTML]{FFFFFF} 
\multicolumn{1}{|l|}{\cellcolor[HTML]{FFFFFF}{\color[HTML]{000000} SACrA}} & \multicolumn{1}{c|}{\cellcolor[HTML]{FFFFFF}{\color[HTML]{000000} \textbf{23.83$^\uparrow$}}} & \multicolumn{1}{c|}{\cellcolor[HTML]{FFFFFF}{\color[HTML]{000000} 15.15}} & \multicolumn{1}{c|}{\cellcolor[HTML]{FFFFFF}{\color[HTML]{000000} 22.86}} & \multicolumn{1}{c|}{\cellcolor[HTML]{FFFFFF}{\color[HTML]{000000} \textbf{18.09$^\uparrow$}}} & \multicolumn{1}{c|}{\cellcolor[HTML]{FFFFFF}{\color[HTML]{000000} 18.13}} & \multicolumn{1}{c|}{\cellcolor[HTML]{FFFFFF}{\color[HTML]{000000} 19.98$^\uparrow$}} & \multicolumn{1}{c|}{\cellcolor[HTML]{FFFFFF}{\color[HTML]{000000} \textbf{18.7$^\uparrow$}}} & \multicolumn{1}{c|}{\cellcolor[HTML]{FFFFFF}{\color[HTML]{000000} 17.1}} & \multicolumn{1}{c|}{\cellcolor[HTML]{FFFFFF}{\color[HTML]{000000} \textbf{13.83$^\uparrow$}}} & \multicolumn{1}{c|}{\cellcolor[HTML]{FFFFFF}{\color[HTML]{000000} 19.41}} & \multicolumn{1}{c|}{\cellcolor[HTML]{FFFFFF}{\color[HTML]{000000} 18.71$^\uparrow$}} \\ \midrule
\multicolumn{1}{|l|}{{\color[HTML]{000000} SACrA + UDISCAL}} & \multicolumn{1}{c|}{\cellcolor[HTML]{FFFFFF}{\color[HTML]{000000} 22.98}} & \multicolumn{1}{c|}{\cellcolor[HTML]{FFFFFF}{\color[HTML]{000000} 14.58}} & \multicolumn{1}{c|}{\cellcolor[HTML]{FFFFFF}{\color[HTML]{000000} 23.16}} & \multicolumn{1}{c|}{\cellcolor[HTML]{FFFFFF}{\color[HTML]{000000} 16.76}} & \multicolumn{1}{c|}{\cellcolor[HTML]{FFFFFF}{\color[HTML]{000000} 17.37}} & \multicolumn{1}{c|}{\cellcolor[HTML]{FFFFFF}{\color[HTML]{000000} 18.89}} & \multicolumn{1}{c|}{\cellcolor[HTML]{FFFFFF}{\color[HTML]{000000} 17.4}} & \multicolumn{1}{c|}{\cellcolor[HTML]{FFFFFF}{\color[HTML]{000000} 16.07}} & \multicolumn{1}{c|}{\cellcolor[HTML]{FFFFFF}{\color[HTML]{000000} 13.18}} & \multicolumn{1}{c|}{{\color[HTML]{000000} 18.53}} & \multicolumn{1}{c|}{{\color[HTML]{000000} 17.89}} \\ \midrule
\multicolumn{9}{c}{\cellcolor[HTML]{FFFFFF}{\color[HTML]{000000} {\ul \textbf{En-Fi}}}} & \cellcolor[HTML]{FFFFFF}{\color[HTML]{000000} {\ul \textbf{}}} & {\color[HTML]{000000} } & {\color[HTML]{000000} } \\ \cmidrule(r){1-9}
\multicolumn{1}{|l|}{{\color[HTML]{000000} \textbf{models}}} & \multicolumn{1}{c|}{{\color[HTML]{000000} \textbf{2015}}} & \multicolumn{1}{c|}{{\color[HTML]{000000} \textbf{2016}}} & \multicolumn{1}{c|}{{\color[HTML]{000000} \textbf{2016B}}} & \multicolumn{1}{c|}{{\color[HTML]{000000} \textbf{2017}}} & \multicolumn{1}{c|}{{\color[HTML]{000000} \textbf{2017B}}} & \multicolumn{1}{c|}{{\color[HTML]{000000} \textbf{2018}}} & \multicolumn{1}{c|}{{\color[HTML]{000000} \textbf{2019}}} & \multicolumn{1}{c|}{{\color[HTML]{000000} {\ul \textbf{average}}}} & {\color[HTML]{000000} {\ul \textbf{}}} & {\color[HTML]{000000} } & {\color[HTML]{000000} } \\ \cmidrule(r){1-9}
\multicolumn{1}{|l|}{{\color[HTML]{000000} Transformer}} & \multicolumn{1}{c|}{{\color[HTML]{000000} 9.57}} & \multicolumn{1}{c|}{{\color[HTML]{000000} \textbf{11.05}}} & \multicolumn{1}{c|}{{\color[HTML]{000000} 8.8}} & \multicolumn{1}{c|}{{\color[HTML]{000000} 11.45}} & \multicolumn{1}{c|}{{\color[HTML]{000000} 9.99}} & \multicolumn{1}{c|}{{\color[HTML]{000000} 7.78}} & \multicolumn{1}{c|}{{\color[HTML]{000000} 10.22}} & \multicolumn{1}{c|}{{\color[HTML]{000000} 9.84}} & {\color[HTML]{000000} } & {\color[HTML]{000000} } & {\color[HTML]{000000} } \\ \cmidrule(r){1-9}
\multicolumn{1}{|l|}{{\color[HTML]{000000} PASCAL}} & \multicolumn{1}{c|}{{\color[HTML]{000000} 9.75}} & \multicolumn{1}{c|}{{\color[HTML]{000000} 10.77}} & \multicolumn{1}{c|}{{\color[HTML]{000000} 8.72}} & \multicolumn{1}{c|}{{\color[HTML]{000000} 11.43}} & \multicolumn{1}{c|}{{\color[HTML]{000000} 10.11}} & \multicolumn{1}{c|}{{\color[HTML]{000000} 8.06}} & \multicolumn{1}{c|}{{\color[HTML]{000000} 10.24}} & \multicolumn{1}{c|}{{\color[HTML]{000000} 9.87}} & {\color[HTML]{000000} \textbf{}} & {\color[HTML]{000000} } & {\color[HTML]{000000} } \\ \cmidrule(r){1-9}
\multicolumn{1}{|l|}{{\color[HTML]{000000} UDISCAL}} & \multicolumn{1}{c|}{{\color[HTML]{000000} 9.04}} & \multicolumn{1}{c|}{{\color[HTML]{000000} 10.85}} & \multicolumn{1}{c|}{{\color[HTML]{000000} 8.63}} & \multicolumn{1}{c|}{{\color[HTML]{000000} 11.46}} & \multicolumn{1}{c|}{{\color[HTML]{000000} 10.1}} & \multicolumn{1}{c|}{{\color[HTML]{000000} 7.7}} & \multicolumn{1}{c|}{{\color[HTML]{000000} 9.85}} & \multicolumn{1}{c|}{{\color[HTML]{000000} 9.66}} & {\color[HTML]{000000} } & {\color[HTML]{000000} } & {\color[HTML]{000000} } \\ \cmidrule(r){1-9}
\multicolumn{1}{|l|}{{\color[HTML]{000000} SASA}} & \multicolumn{1}{c|}{\cellcolor[HTML]{FFFFFF}{\color[HTML]{000000} 9.65}} & \multicolumn{1}{c|}{\cellcolor[HTML]{FFFFFF}{\color[HTML]{000000} 10.87}} & \multicolumn{1}{c|}{\cellcolor[HTML]{FFFFFF}{\color[HTML]{000000} \textbf{9.03$^\uparrow$}}} & \multicolumn{1}{c|}{\cellcolor[HTML]{FFFFFF}{\color[HTML]{000000} 11.62$^\uparrow$}} & \multicolumn{1}{c|}{\cellcolor[HTML]{FFFFFF}{\color[HTML]{000000} 10.1}} & \multicolumn{1}{c|}{\cellcolor[HTML]{FFFFFF}{\color[HTML]{000000} 7.99}} & \multicolumn{1}{c|}{\cellcolor[HTML]{FFFFFF}{\color[HTML]{000000} 10.53$^\uparrow$}} & \multicolumn{1}{c|}{\cellcolor[HTML]{FFFFFF}{\color[HTML]{000000} 9.97$^\uparrow$}} & \cellcolor[HTML]{FFFFFF}{\color[HTML]{000000} \textbf{}} & {\color[HTML]{000000} } & {\color[HTML]{000000} } \\ \cmidrule(r){1-9}
\multicolumn{1}{|l|}{{\color[HTML]{000000} SASA + UDISCAL}} & \multicolumn{1}{c|}{\cellcolor[HTML]{FFFFFF}{\color[HTML]{000000} 9.45}} & \multicolumn{1}{c|}{\cellcolor[HTML]{FFFFFF}{\color[HTML]{000000} 10.96$^{*}$}} & \multicolumn{1}{c|}{\cellcolor[HTML]{FFFFFF}{\color[HTML]{000000} 8.91}} & \multicolumn{1}{c|}{\cellcolor[HTML]{FFFFFF}{\color[HTML]{000000} \textbf{11.88$^{*}$}}} & \multicolumn{1}{c|}{\cellcolor[HTML]{FFFFFF}{\color[HTML]{000000} \textbf{10.33$^{*}$}}} & \multicolumn{1}{c|}{\cellcolor[HTML]{FFFFFF}{\color[HTML]{000000} \textbf{8.42$^{*}$}}} & \multicolumn{1}{c|}{\cellcolor[HTML]{FFFFFF}{\color[HTML]{000000} 10.62$^{*}$}} & \multicolumn{1}{c|}{\cellcolor[HTML]{FFFFFF}{\color[HTML]{000000} 10.08$^{*}$}} & {\color[HTML]{000000} \textbf{}} & {\color[HTML]{000000} } & {\color[HTML]{000000} } \\ \cmidrule(r){1-9}
\multicolumn{1}{|l|}{{\color[HTML]{000000} SACrA}} & \multicolumn{1}{c|}{\cellcolor[HTML]{FFFFFF}{\color[HTML]{000000} \textbf{10.26$^\uparrow$}}} & \multicolumn{1}{c|}{\cellcolor[HTML]{FFFFFF}{\color[HTML]{000000} 10.95}} & \multicolumn{1}{c|}{\cellcolor[HTML]{FFFFFF}{\color[HTML]{000000} 8.89$^\uparrow$}} & \multicolumn{1}{c|}{\cellcolor[HTML]{FFFFFF}{\color[HTML]{000000} 11.57$^\uparrow$}} & \multicolumn{1}{c|}{\cellcolor[HTML]{FFFFFF}{\color[HTML]{000000} 10.13$^\uparrow$}} & \multicolumn{1}{c|}{\cellcolor[HTML]{FFFFFF}{\color[HTML]{000000} 8.17$^\uparrow$}} & \multicolumn{1}{c|}{\cellcolor[HTML]{FFFFFF}{\color[HTML]{000000} \textbf{10.76$^\uparrow$}}} & \multicolumn{1}{c|}{{\color[HTML]{000000} \textbf{10.10$^\uparrow$}}} & {\color[HTML]{000000} \textbf{}} & {\color[HTML]{000000} } & {\color[HTML]{000000} } \\ \cmidrule(r){1-9}
\multicolumn{1}{|l|}{{\color[HTML]{000000} SACrA + UDISCAL}} & \multicolumn{1}{c|}{{\color[HTML]{000000} 9.42}} & \multicolumn{1}{c|}{{\color[HTML]{000000} 10.84}} & \multicolumn{1}{c|}{{\color[HTML]{000000} 8.83}} & \multicolumn{1}{c|}{{\color[HTML]{000000} 11.51}} & \multicolumn{1}{c|}{{\color[HTML]{000000} 9.9}} & \multicolumn{1}{c|}{{\color[HTML]{000000} 7.71}} & \multicolumn{1}{c|}{{\color[HTML]{000000} 10.7}} & \multicolumn{1}{c|}{{\color[HTML]{000000} 9.84}} & {\color[HTML]{000000} } & {\color[HTML]{000000} } & {\color[HTML]{000000} } \\ \cmidrule(r){1-9}
\rowcolor[HTML]{FFFFFF} 
\multicolumn{9}{c}{\cellcolor[HTML]{FFFFFF}{\color[HTML]{000000} {\ul \textbf{En-Tr}}}} & {\color[HTML]{000000} {\ul \textbf{}}} & {\color[HTML]{000000} } & {\color[HTML]{000000} } \\ \cmidrule(r){1-9}
\multicolumn{1}{|l|}{{\color[HTML]{000000} \textbf{models}}} & \multicolumn{1}{c|}{{\color[HTML]{000000} \textbf{2016}}} & \multicolumn{1}{c|}{{\color[HTML]{000000} \textbf{2017}}} & \multicolumn{1}{c|}{{\color[HTML]{000000} \textbf{2018}}} & \multicolumn{1}{c|}{{\color[HTML]{000000} \textbf{wikipedia}}} & \multicolumn{1}{c|}{{\color[HTML]{000000} \textbf{Eubookshop}}} & \multicolumn{1}{c|}{{\color[HTML]{000000} \textbf{mozilla}}} & \multicolumn{1}{c|}{{\color[HTML]{000000} \textbf{bible}}} & \multicolumn{1}{c|}{{\color[HTML]{000000} {\ul \textbf{average}}}} & {\color[HTML]{000000} {\ul \textbf{}}} & {\color[HTML]{000000} } & {\color[HTML]{000000} } \\ \cmidrule(r){1-9}
\multicolumn{1}{|l|}{{\color[HTML]{000000} Transformer}} & \multicolumn{1}{c|}{\cellcolor[HTML]{FFFFFF}{\color[HTML]{000000} 7.99}} & \multicolumn{1}{c|}{\cellcolor[HTML]{FFFFFF}{\color[HTML]{000000} 8.15}} & \multicolumn{1}{c|}{\cellcolor[HTML]{FFFFFF}{\color[HTML]{000000} 8.06}} & \multicolumn{1}{c|}{\cellcolor[HTML]{FFFFFF}{\color[HTML]{000000} 7.55}} & \multicolumn{1}{c|}{\cellcolor[HTML]{FFFFFF}{\color[HTML]{000000} 4.87}} & \multicolumn{1}{c|}{\cellcolor[HTML]{FFFFFF}{\color[HTML]{000000} 3.34}} & \multicolumn{1}{c|}{\cellcolor[HTML]{FFFFFF}{\color[HTML]{000000} 0.36}} & \multicolumn{1}{c|}{\cellcolor[HTML]{FFFFFF}{\color[HTML]{000000} 5.76}} & {\color[HTML]{000000} } & {\color[HTML]{000000} } & {\color[HTML]{000000} } \\ \cmidrule(r){1-9}
\multicolumn{1}{|l|}{{\color[HTML]{000000} PASCAL}} & \multicolumn{1}{c|}{\cellcolor[HTML]{FFFFFF}{\color[HTML]{000000} 7.81}} & \multicolumn{1}{c|}{\cellcolor[HTML]{FFFFFF}{\color[HTML]{000000} 7.83}} & \multicolumn{1}{c|}{\cellcolor[HTML]{FFFFFF}{\color[HTML]{000000} 7.69}} & \multicolumn{1}{c|}{\cellcolor[HTML]{FFFFFF}{\color[HTML]{000000} 7.52}} & \multicolumn{1}{c|}{\cellcolor[HTML]{FFFFFF}{\color[HTML]{000000} 5.04}} & \multicolumn{1}{c|}{\cellcolor[HTML]{FFFFFF}{\color[HTML]{000000} 3.41}} & \multicolumn{1}{c|}{\cellcolor[HTML]{FFFFFF}{\color[HTML]{000000} \textbf{0.54}}} & \multicolumn{1}{c|}{\cellcolor[HTML]{FFFFFF}{\color[HTML]{000000} 5.69}} & {\color[HTML]{000000} \textbf{}} & {\color[HTML]{000000} } & {\color[HTML]{000000} } \\ \cmidrule(r){1-9}
\multicolumn{1}{|l|}{{\color[HTML]{000000} UDISCAL}} & \multicolumn{1}{c|}{\cellcolor[HTML]{FFFFFF}{\color[HTML]{000000} 7.68}} & \multicolumn{1}{c|}{\cellcolor[HTML]{FFFFFF}{\color[HTML]{000000} 7.83}} & \multicolumn{1}{c|}{\cellcolor[HTML]{FFFFFF}{\color[HTML]{000000} 7.4}} & \multicolumn{1}{c|}{\cellcolor[HTML]{FFFFFF}{\color[HTML]{000000} 7.63}} & \multicolumn{1}{c|}{\cellcolor[HTML]{FFFFFF}{\color[HTML]{000000} 4.92}} & \multicolumn{1}{c|}{\cellcolor[HTML]{FFFFFF}{\color[HTML]{000000} 3.34}} & \multicolumn{1}{c|}{\cellcolor[HTML]{FFFFFF}{\color[HTML]{000000} 0.49}} & \multicolumn{1}{c|}{\cellcolor[HTML]{FFFFFF}{\color[HTML]{000000} 5.61}} & {\color[HTML]{000000} \textbf{}} & {\color[HTML]{000000} } & {\color[HTML]{000000} } \\ \cmidrule(r){1-9}
\multicolumn{1}{|l|}{{\color[HTML]{000000} SASA}} & \multicolumn{1}{c|}{\cellcolor[HTML]{FFFFFF}{\color[HTML]{000000} 8.2$^\uparrow$}} & \multicolumn{1}{c|}{\cellcolor[HTML]{FFFFFF}{\color[HTML]{000000} 8.31$^\uparrow$}} & \multicolumn{1}{c|}{\cellcolor[HTML]{FFFFFF}{\color[HTML]{000000} \textbf{8.12$^\uparrow$}}} & \multicolumn{1}{c|}{\cellcolor[HTML]{FFFFFF}{\color[HTML]{000000} 7.63}} & \multicolumn{1}{c|}{\cellcolor[HTML]{FFFFFF}{\color[HTML]{000000} 5.21$^\uparrow$}} & \multicolumn{1}{c|}{\cellcolor[HTML]{FFFFFF}{\color[HTML]{000000} 3.09}} & \multicolumn{1}{c|}{\cellcolor[HTML]{FFFFFF}{\color[HTML]{000000} 0.52}} & \multicolumn{1}{c|}{\cellcolor[HTML]{FFFFFF}{\color[HTML]{000000} 5.87$^\uparrow$}} & {\color[HTML]{000000} \textbf{}} & {\color[HTML]{000000} } & {\color[HTML]{000000} } \\ \cmidrule(r){1-9}
\multicolumn{1}{|l|}{{\color[HTML]{000000} SASA + UDISCAL}} & \multicolumn{1}{c|}{\cellcolor[HTML]{FFFFFF}{\color[HTML]{000000} 7.81}} & \multicolumn{1}{c|}{\cellcolor[HTML]{FFFFFF}{\color[HTML]{000000} 7.92}} & \multicolumn{1}{c|}{\cellcolor[HTML]{FFFFFF}{\color[HTML]{000000} 8.1}} & \multicolumn{1}{c|}{\cellcolor[HTML]{FFFFFF}{\color[HTML]{000000} 7.58}} & \multicolumn{1}{c|}{\cellcolor[HTML]{FFFFFF}{\color[HTML]{000000} \textbf{5.28$^{*}$}}} & \multicolumn{1}{c|}{\cellcolor[HTML]{FFFFFF}{\color[HTML]{000000} 3.36$^{*}$}} & \multicolumn{1}{c|}{\cellcolor[HTML]{FFFFFF}{\color[HTML]{000000} 0.35}} & \multicolumn{1}{c|}{\cellcolor[HTML]{FFFFFF}{\color[HTML]{000000} 5.77}} & {\color[HTML]{000000} \textbf{}} & {\color[HTML]{000000} } & {\color[HTML]{000000} } \\ \cmidrule(r){1-9}
\multicolumn{1}{|l|}{{\color[HTML]{000000} SACrA}} & \multicolumn{1}{c|}{\cellcolor[HTML]{FFFFFF}{\color[HTML]{000000} 7.75}} & \multicolumn{1}{c|}{\cellcolor[HTML]{FFFFFF}{\color[HTML]{000000} 8.33$^\uparrow$}} & \multicolumn{1}{c|}{\cellcolor[HTML]{FFFFFF}{\color[HTML]{000000} 7.51}} & \multicolumn{1}{c|}{\cellcolor[HTML]{FFFFFF}{\color[HTML]{000000} \textbf{7.68$^\uparrow$}}} & \multicolumn{1}{c|}{\cellcolor[HTML]{FFFFFF}{\color[HTML]{000000} 5.11$^\uparrow$}} & \multicolumn{1}{c|}{\cellcolor[HTML]{FFFFFF}{\color[HTML]{000000} \textbf{3.59$^\uparrow$}}} & \multicolumn{1}{c|}{\cellcolor[HTML]{FFFFFF}{\color[HTML]{000000} 0.5}} & \multicolumn{1}{c|}{\cellcolor[HTML]{FFFFFF}{\color[HTML]{000000} 5.78$^\uparrow$}} & {\color[HTML]{000000} \textbf{}} & {\color[HTML]{000000} } & {\color[HTML]{000000} } \\ \cmidrule(r){1-9}
\multicolumn{1}{|l|}{{\color[HTML]{000000} SACrA + UDISCAL}} & \multicolumn{1}{c|}{\cellcolor[HTML]{FFFFFF}{\color[HTML]{000000} \textbf{8.23$^{*}$}}} & \multicolumn{1}{c|}{\cellcolor[HTML]{FFFFFF}{\color[HTML]{000000} \textbf{8.54$^{*}$}}} & \multicolumn{1}{c|}{\cellcolor[HTML]{FFFFFF}{\color[HTML]{000000} 7.95$^{*}$}} & \multicolumn{1}{c|}{\cellcolor[HTML]{FFFFFF}{\color[HTML]{000000} 7.51}} & \multicolumn{1}{c|}{\cellcolor[HTML]{FFFFFF}{\color[HTML]{000000} 5.22$^{*}$}} & \multicolumn{1}{c|}{\cellcolor[HTML]{FFFFFF}{\color[HTML]{000000} 3.45}} & \multicolumn{1}{c|}{\cellcolor[HTML]{FFFFFF}{\color[HTML]{000000} 0.52$^{*}$}} & \multicolumn{1}{c|}{\cellcolor[HTML]{FFFFFF}{\color[HTML]{000000} \textbf{5.92$^{*}$}}} & {\color[HTML]{000000} \textbf{}} & {\color[HTML]{000000} } & {\color[HTML]{000000} } \\ \cmidrule(r){1-9}
\end{tabular}%
}
\caption{Bleu scores of challenge sentences for the baseline Transformer model, the baseline Syntactically infused models PASCAL and UDISCAL, our SASA and SACrA models, and models incorporating UDISCAL with each of SASA and SACrA, across all WMT's newstests. For every language pair, each column contains the Bleu scores over the WMT newstest equivalent to the column's year (e.g., for En-Ru, the scores under column \textit{2015} are for En-Ru newstest2015). For some newstests, there was more than one version on WMT, each translated by a different person. For those test sets, we included both versions, denoting the second one with a "B". In addition, for every language pair, the right-most column represents the average Bleu scores over all the pair's reported newstests. For every test set (and for the average score), the best score is boldfaced. For each of the semantic models (i.e., SASA and SACrA), improvements over all the baselines (syntactic and Transformer) are marked by an arrow facing upwards. For models with both syntactic and semantic masks, improvements over each mask individually are marked by an asterisk.}
\label{tab:challenge_sets_Bleu}
\end{table*}
    
    \subsection{SemSplit}\label{subsec:SemSplit}
        Following \citet{sulem-etal-2020-semantic}, we implement the SemSplit pipeline. First, we train a Transformer-based Neural Machine Translation model. Then, during inference time, we use the Direct Semantic Splitting algorithm \citep[DSS;][]{sulem-etal-2018-simple} to split the sentences, and then translate each separated sentence separately. Finally, we concatenate the translation, using a period (".") as a delimiter. Table \ref{tab:semSplit_results} presents the results, using the Bleu and chrF metrics. We find that the architecture does not have gains over the baseline Transformer. These results can be accounted for by the fact that in their work, \citet{sulem-etal-2020-semantic} assessed the pipeline's performance using Human Evaluation and manual analysis, rather than the Bleu and chrF metrics, which punish for sentence separation in translation. In addition, they tested their pipeline in a pseudo-low resource scenario, and not in normal NMT settings.

\begin{figure*}
     \centering
     \begin{subfigure}[b]{0.4\textwidth}
         \centering
         \includegraphics[width=\textwidth]{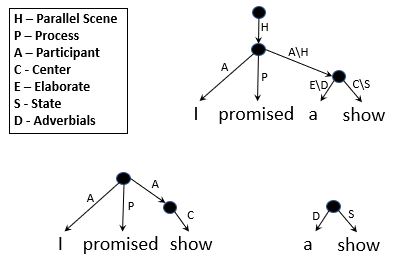}
         \caption{I promised a show?}
         \label{fig:Qualitative_Analysis_example1}
     \end{subfigure}
     \hfill
     \begin{subfigure}[b]{0.55\textwidth}
         \centering
         \includegraphics[width=\textwidth]{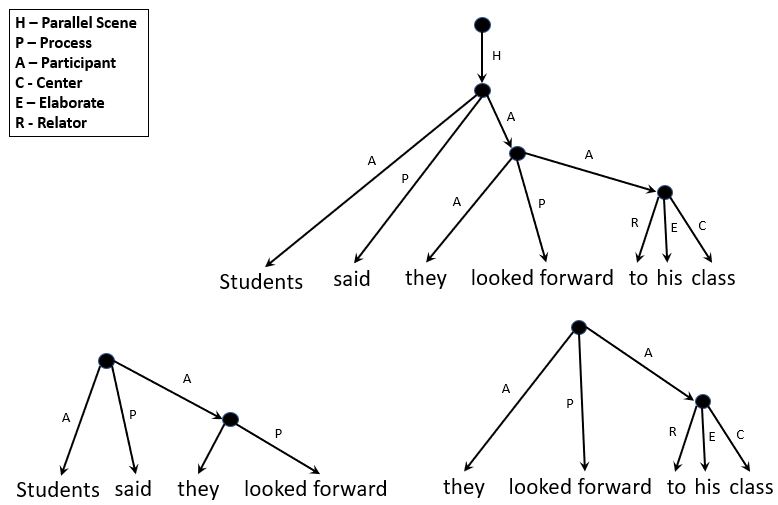}
         \caption{Students said they looked forward to his class.}
         \label{fig:Qualitative_Analysis_example2}
     \end{subfigure}
     \hfill
     \begin{subfigure}[b]{0.8\textwidth}
         \centering
         \includegraphics[width=\textwidth]{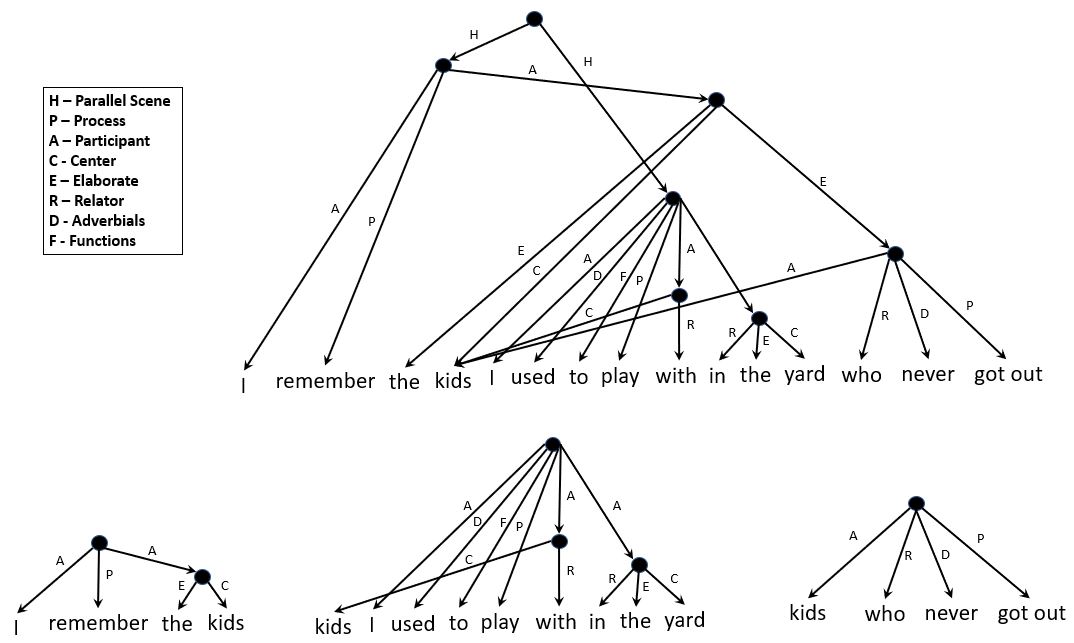}
         \caption{I remember those kids I used to play with in the yard who never got out.}
         \label{Qualitative_Analysis_example3}
     \end{subfigure}
        \caption{UCCA parse graphs of the Qualitative Analysis examples, with the equivalent UCCA sub-graphs representing the segmentation into scenes.}
        \label{fig:Qualitative_Analysis_examples}
\end{figure*}

    \subsection{Qualitative Analysis - UCCA Parsings}\label{subsec:Qualitative Analysis - UCCA Parsings}
    figure \ref{fig:Qualitative_Analysis_examples} presents the UCCA parsings of the examples featured in table \ref{tab:SASA_vs_base examples}.
    
\end{document}